\documentclass[5p,twocolumn]{elsarticle}

\usepackage{amssymb}

\journal{Information Sciences}

\usepackage{multicol}
\usepackage{amsmath,mathrsfs}
\usepackage{algorithm}
\usepackage{algorithmicx}
\usepackage{color}
\usepackage{algpseudocode}

\usepackage{CJKutf8}
\usepackage{subfigure}
\usepackage{verbatim}
\usepackage{multirow}
\usepackage{graphicx}
\usepackage{subfigure}
\usepackage{hyperref}
\usepackage{booktabs}
 \usepackage{threeparttable}

\usepackage{xparse}%

\usepackage{blindtext}

\usepackage{morewrites}

\usepackage{wrapfig}
\usepackage{xkeyval}%

\newwrite\authorbibfile%

\AtBeginDocument{%
	\immediate\openout\authorbibfile=\jobname.aub%
}%
\AtEndDocument{%
	\immediate\closeout\authorbibfile
	\InputIfFileExists{\jobname.aub}{}{}
}%

\makeatletter

\define@key{authorbib}{scale}[1]{%
	\def\AuthorbibKVMacroScale{#1}%
}

\define@key{authorbib}{wraplines}[10]{%
	\def\AuthorbibKVMacroWraplines{#1}%
}

\define@key{authorbib}{imagewidth}[4cm]{%
	\def\AuthorbibKVMacroImagewidth{#1}%
}
\define@key{authorbib}{overhang}[10pt]{%
	\def\AuthorbibKVMacroOverhang{#1}%
}

\define@key{authorbib}{imagepos}[l]{%
	\def\AuthorbibKVMacroImagepos{#1}%
}

\makeatother

\presetkeys{authorbib}{imagepos=l,imagewidth=4cm,wraplines=15,overhang=20pt}{}

\newlength{\AuthorbibTopSkip}
\newlength{\AuthorbibBottomSkip}
\NewDocumentCommand{\authorbibliography}{+o+m+m+m}{%
	\IfNoValueTF{#1}{%
	}{%
		\setkeys{authorbib}{#1}%
		\immediate\write\authorbibfile{%
			\string\begin{wrapfigure}[\AuthorbibKVMacroWraplines]{\AuthorbibKVMacroImagepos}[\AuthorbibKVMacroOverhang]{\AuthorbibKVMacroImagewidth}^^J
				\string\includegraphics[scale=\AuthorbibKVMacroScale]{#2}^^J
				\string\end{wrapfigure}^^J
		}%
	}%
	\IfNoValueTF{#3}{%
		\typeout{Warning: No author name}%
	}{%
		\immediate\write\authorbibfile{%
			\unexpanded{\vspace{\AuthorbibTopSkip}}^^J
			\string\noindent\relax
			\unexpanded{\textbf{#3}\par}^^J
			\string\noindent\relax
			\unexpanded{#4}^^J%
			\unexpanded{\vspace{\AuthorbibBottomSkip}}^^J
		}%
	}%
}%

\begin{document}

\begin{frontmatter}



\title{Exploiting Style Transfer-based Task Augmentation \\for Cross-Domain Few-Shot Learning}


\author{Shuzhen Rao}
\author{Jun Huang \corref{cor}}
\author{Zengming Tang}

\address{Shanghai Advanced Research Institute, Chinese Academy of Sciences, Shanghai, 201210, PR China}
\address{University of Chinese Academy of Sciences, Beijing, 100049, PR China}
        
\cortext[cor]{Jun Huang is the corresponding author.\\ \indent \indent E-mail address: huangj@sari.ac.cn.}

\begin{abstract}
	In cross-domain few-shot learning, the core issue is that the model trained on source domains struggles to generalize to the target domain, especially when the domain shift is large.
	Motivated by the observation that the domain shift between training tasks and target tasks usually can reflect in their style variation, we propose Task Augmented Meta-Learning (TAML) to conduct style transfer-based task augmentation to improve the domain generalization ability.
	Firstly, Multi-task Interpolation (MTI) is introduced to fuse features from multiple tasks with different styles, which makes more diverse styles available. Furthermore, a novel task-augmentation strategy called Multi-Task Style Transfer (MTST) is proposed to perform style transfer on existing tasks to learn discriminative style-independent features. 
	We also introduce a Feature Modulation module (FM) to add random styles and improve generalization of the model.
	The proposed TAML increases the diversity of styles of training tasks, and contributes to training a model with better domain generalization ability.
	The effectiveness is demonstrated via theoretical analysis and thorough experiments on two popular cross-domain few-shot benchmarks.
\end{abstract}

%


\end{frontmatter}



\section{Introduction}

Few-shot learning (FSL) aims to classify query samples from novel classes with only few labeled support samples in each class. 
Recently, various approaches have been proposed to addressing the FSL problem \cite{snell2017prototypical,sung2018learning,vinyals2016matching,finn2017model,hu2020learning,zhu2020attribute,huang2020low,liu2022adaptive,li2022coarse,qin2022multi}.
Among these FSL methods, metric-based meta-learning methods \cite{snell2017prototypical,vinyals2016matching,sung2018learning,huang2020low,li2022coarse,qin2022multi} have achieved impressive performance when testing tasks are sampled from the same domain as training tasks. 
In general, they make the prediction based on the similarity between features of query samples and support samples.
However, these methods can not generalize well to novel classes from unseen domains. In some practical application scenarios, constructing large training datasets for rare classes is almost impossible, making it essential to improve the generalization ability of the model to unseen domains.
As a result, the cross-domain few-shot learning (CD-FSL) problem, where training tasks and target tasks are sampled from completely different domains, has received considerable attention \cite{tseng2020cross}. The core issue in CD-FSL is how to understand and address the large domain shift between source domains and the target domain. If the domain shift is very large, the feature extractor and metric function trained on source tasks sampled from source domains can not generalize well to target tasks sampled from target domain, especially when there are only a few samples in each task.
\begin{center}
	\begin{figure}[t]
		\centering
		\includegraphics[width=\linewidth]{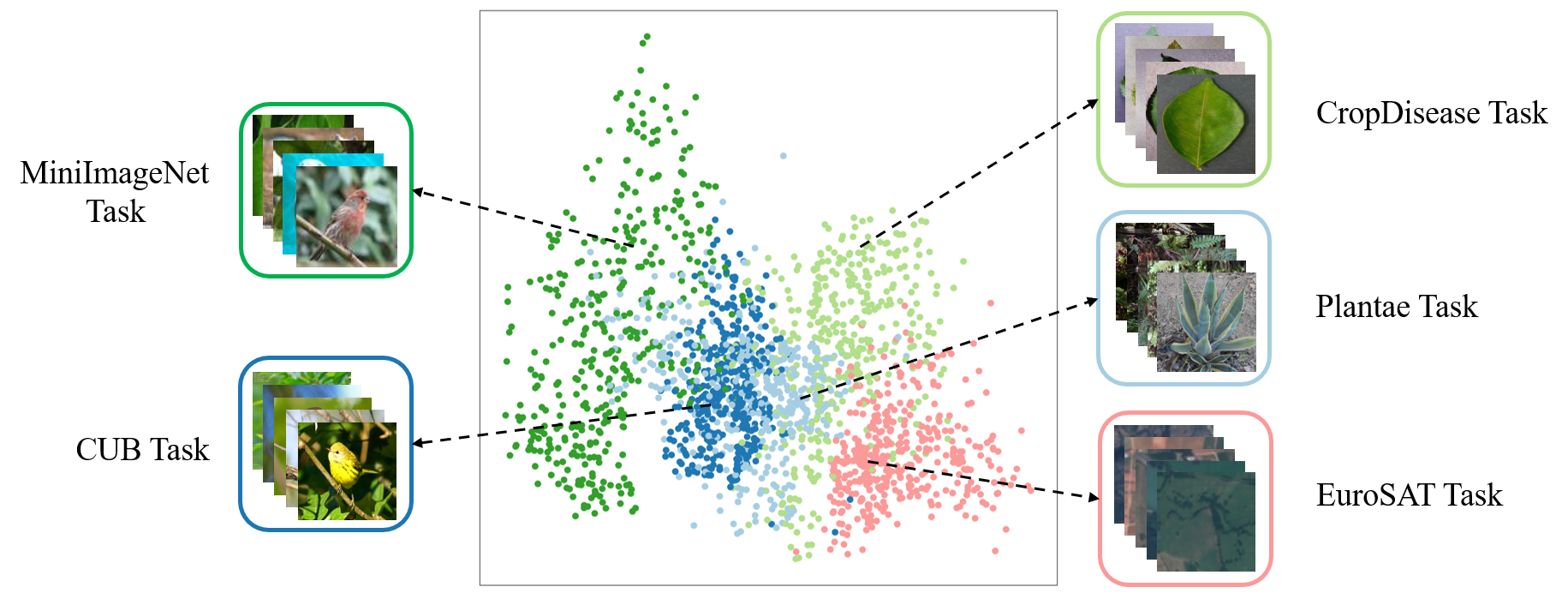}
		\caption{T-SNE visualization of the task style statistics. Each node refers to the concatenation of mean and variance of the correspondingly task features, which are calculated with a random residual block of pretrained ResNet-10. Style variation exists in training tasks and target tasks.}
		\label{fig:taskstyle}
	\end{figure}
\end{center}
\vspace{-0.8cm}

Cross-domain few-shot learning methods \cite{tseng2020cross,sun2021explanation,fu2021meta,wang2021cross} emphasize that target tasks contain novel categories from unseen domains. It is observed that the domain shift between training tasks and target tasks can reflect in differences in their styles, including resolution, color contrast, and illumination \cite{jin2021style}. The style of an image can be quantified by the mean and variance of its features \cite{zhou2021domain}. Similarly, in our task-level style representation, the mean and variance of all image features within a given task are used. We visualize the task style statistics of different domains in Figure \ref{fig:taskstyle} to analyze the differences between training tasks and target tasks.
It is evident in this figure that, style variation exists between training tasks and target tasks. A model with good generalization ability should extract discriminative domain-independent or style-independent features.
As demonstrated by \cite{zhang2021understanding,jin2021style}, increasing the diversity of the dataset can help train a style-independent model, substantially improving the generalization of representations.
Some cross-domain few-shot learning methods, such as ATA \cite{wang2021cross} and AFA \cite{hu2022adversarial}, have attempted to implicitly increase the style variety of training tasks and improve generalization by adversarial-based augmentation. However, these methods can be limited in effectiveness, particularly when there are only a few samples available for training in CD-FSL, as they are computationally intensive and require careful hyperparameter tuning.

To address these challenges, we propose Task-Augmented Meta-Learning (TAML), which expands the diversity of source domains by using style transfer-based task augmentation. This is the first work to use style transfer for task augmentation.
Our method uses Multi-Task Interpolation (MTI) to perform feature fusion on multiple original tasks from source domains and generate an interpolated task. MTI allows for training with more diverse domains and styles in a more efficient and effective way than pairwise task interpolation.
We then introduce Multi-Task Style Transfer (MTST) to learn a style-independent model using more diverse styles of training tasks. With the interpolated task, we obtain new task style parameters and use them to perform style transfer on original tasks to generate style-transferred tasks. By leveraging the original class information, training on style-transferred tasks enables the model to extract style-independent features and achieve better domain generalization.
To simulate more feature distributions by adding random styles, we use the Feature Modulation module (FM) to meta-learn to transform features of new tasks generated by MTST.
Our proposed task augmentation method effectively increases the diversity of training task styles, leading to a more style-independent and domain-generalizable model.

In summary, we make the following contributions in this work:

\begin{itemize}
	\item We propose a novel style transfer-based task augmentation method called Task-Augmented Meta-Learning (TAML) to bridge the domain shift in CD-FSL through more styles of training tasks. 
	
	\item Multi-task Interpolation (MTI) is introduced to perform feature fusion on any number of tasks efficiently and effectively, which increases the diversity of training tasks and makes more diverse styles available. 
	
	\item A novel task-augmentation strategy called Multi-Task Style Transfer (MTST) is put forward to perform style transfer on existing tasks based on styles obtained by MTI. MTST further utilizes original category information to learn discriminative style-independent features. 
	
	\item We introduce Feature Modulation module (FM) to affine transform styles of new tasks, thereby simulating various styles and training a more domain-generalizable model.
	
	\item TAML empirically outperforms state-of-the-art CD-FSL approaches on two popular cross-domain few-shot benchmarks, demonstrating its significant effectiveness.
\end{itemize}




\section{Related work}
\subsection{Few-shot learning}
In the Few-shot Learning (FSL) problem, only a limited number of samples with supervised information are available. Existing few-shot learning methods can be roughly categorized into two groups, optimization-based methods \cite{finn2017model,antoniou2019how} and metric-based methods \cite{vinyals2016matching,sung2018learning}. Optimization-based methods meta-learn a generalizable model initialization, and then adapt the model to a novel task with a few number of SGD steps. Metric-based frameworks embed samples into a low dimensional space, where congeneric samples are closer together while inhomogeneous samples can be easier to be differentiated. 

The above methods assume that samples for training and testing are from the same domain. Research \cite{chen2019closer} points out that existing metric-based few-shot methods fail to generalize to novel target classes when the source dataset and the target dataset are disjoint. Thus, in this paper, we focus on metric-based methods and aim to improve their performance in the CD-FSL scenario where the source and target datasets are disjoint.

\subsection{Cross-domain few-shot learning}
Although various meta-learning models for few-shot classification have achieved impressive performance, they fail to generalize to unseen domains because of domain shift between the source domain and target domain. Cross-domain few-shot learning is a branch of few shot image classification, where training and target tasks are sampled from different domains. We can make use of these data of different domains to make the few-shot classifier more robust. A new benchmark called ECCV 2020 challenge or BSCD-FSL \cite{guo2020broader} has been proposed for this problem.

Some cross-domain few-shot learning methods improve the performance on the target domain by batch spectral regularization, model ensemble and large margin mechanism. The group of methods can cause a huge burden of complexity and memory. Other methods focus on augmenting the meta-training set from different perspectives. 
FWT \cite{tseng2020cross} proposes feature-wise transformation layers to simulate various domains of image features during training.
LRP \cite{sun2021explanation} dynamically finds and emphasizes the features which are important for the predictions based on existing explanation methods. 
Inspired by Xmixup \cite{li2020xmixup}, Meta-FDMixup \cite{fu2021meta} mixes source images and newly introduced auxiliary images, and learns to disentangle more distinguishable domain-irrelevant and the domain-specific images features.
Different from above task-agnostic methods, ATA \cite{wang2021cross} generates the inductive bias-adaptive `challenging’ tasks through adversarial task augmentation to improve the robustness of inductive bias. On the basis of ATA, AFA \cite{hu2022adversarial} proposes to generate augmented features to simulate domain variance. However, the adversarial augmentation is complex and shows limited effect for the few-shot setting. Our method conduct feature-based task augmentation by interpolation and style transfer, which is simpler yet more effective.

\begin{center}
	\begin{figure*}[t]
		\centering
		\includegraphics[width=.8\linewidth]{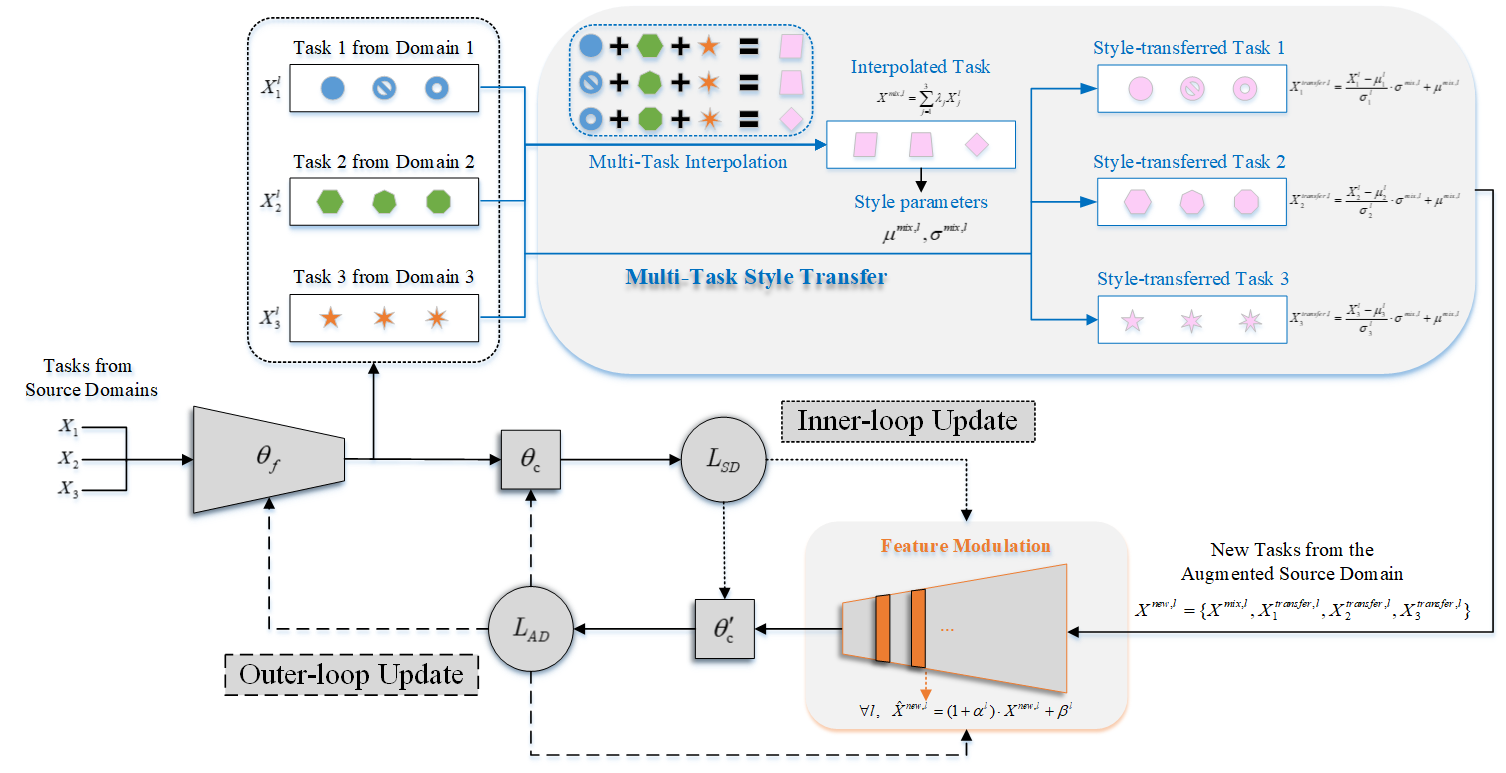}
		\caption{The diagram of our Task-Augmentation Meta-Learning (TAML). The key idea is to conduct style transfer-based task augmentation. Concretely, we propose Multi-Task Interpolation, Multi-Task Style Transfer and Feature Modulation, which augments the source domain by conducting style fusion, style transfer and random style transformation to tasks. The proposed task augmentation increases the diversity of training tasks and improves the generalization ability of the model.}
		\label{fig:diagram}
	\end{figure*}
\end{center}

\subsection{Data augmentation}
An advanced data augmentation method termed as Mixup \cite{zhang2017mixup} is proposed to alleviate the overfitting issue and improve the robustness of the network. On the basis of mixup, some variants have been proposed recently, including CutMix \cite{yun2019cutmix}, Manifold Mixup \cite{verma2019manifold}, AugMix \cite{hendrycks2019augmix}, PuzzleMix \cite{kim2020puzzle} and so on. These mixup methods are designed for the classical image classification task.
Recent domain-agnostic techniques, e.g. Meta-MaxUp \cite{ni2021data} and MetaMix \cite{yao2021improving}, have augmented tasks by applying Mixup and its variants to each task. However, these techniques can not increase the number of meta-training tasks, and will not work in improving the generalization to new tasks.
Unlike these domain-agnostic augmentation strategies that apply data augmentation on each task individually, MLTI \cite{yao2021meta} directly densifies the task distribution by generating additional tasks from pairs of existing tasks. As a task-level data augmentation methods, MLTI outperforms all of these above instance-level data augmentation techniques, because it considers the limited number of tasks and densifies the task distribution for training. 
Thus, we are inspired to conduct task interpolation and style transfer on multiple tasks from source domains, making more styles of training tasks available and the model more generalizable. Different from Mixstyle \cite{zhou2021domain}, we utilize the feature statistics of training tasks to simulate tasks from different domains rather than samples. That is due to the fact that, task-level style parameters are more representative of the overall statistical characteristics of the domain than instances.

\subsection{Domain generalization}

Domain generalization (DG) methods aim to generalize from seen source domains to the unseen target domain as well without using samples from them \cite{khosla2012undoing}. DG methods can be split into three categories, i.e., feature-based methods, metric-based methods and data augmentation methods. The feature-based group of methods learn to extract domain-invariant features across source domains \cite{muandet2013domain}, \cite{li2019feature}. Metric-based methods enhance generalization by adding a manually designed loss function \cite{balaji2018metareg} or fusing multiple sub-classifiers learned from source domains \cite{niu2015multi}. Data augmentation methods augment source domains by generating new samples and utilize them to train a more robust model \cite{li2019feature}. For example, ADA \cite{volpi2018generalizing} is proposed to conduct adaptive data augmentation by appending adversarial examples at each iteration. Our approach simulates domain shift by augmenting source domains, and meta-learns a robust task-shared feature extractor.
%

\section{Proposed Method}
In this section, we introduce the Cross-domain Few-shot Classification problem firstly. Then we illustrate the proposed Task-Augmented Meta-Learning (TAML) and describe the step-by-step algorithm, which consists of the proposed task augmentation and the meta-learning on the augmented source domains. The diagram is shown in Figure \ref{fig:diagram}.

\subsection{Preliminaries}
\subsubsection{The Metric-based Methods for Few-shot Classification}
Assume that the task distribution or domain is $ d_0 $. Each few-shot task $ T_j $ consists of a support set $ \mathcal{S}_j $ and a query set $ \mathcal{Q}_j $, i.e., $ T_j=\left\{\mathcal{S}_j, \mathcal{Q}_j \right\}$, where $  \mathcal{S}_j=\left\{(x_{i,j}^s,y_{i,j}^s)\right\}_{i=1}^{N \times K_s } $ and $  \mathcal{Q}_j=\left\{(x_{i,j}^q,y_{i,j}^q)\right\}_{i=1}^{N \times K_q } $. Metric-based algorithms meta-learn a feature extractor $ F $ and a classifier $ C $, which are parameterized by $ \theta_f $ and $ \theta_c $ respectively. For each task, we extract features of all samples from both $  \mathcal{S} $ and $ \mathcal{Q} $ with $ F $, and then classify samples in $ \mathcal{Q} $ based on $ \mathcal{S} $ with $ C $:

\begin{equation}
	\begin{aligned}
		\hat{y_i^q}= \theta_c(\theta_f(x_i^q)).
	\end{aligned}
\end{equation}

The main difference among meta-learning models for few-shot classification lies in the design choices for the classifier $ \theta_c $. In this paper, we consider the following three different classifiers, i.e., MatchingNet \cite{vinyals2016matching}, RelationNet \cite{sung2018learning} and GNN \cite{satorras2018few}, which are commonly used for performance comparison in CD-FSL. 


\subsubsection{The Cross-domain Few-shot Classification Setting}
In traditional few-shot learning, both training and testing tasks are assumed to be sampled from the same task distribution $ d_0 $. However, in this work, we focus on the cross-domain few-shot classification, where the domains themselves are viewed as distributions of few-shot classification tasks. Specifically, we assume that the training tasks are sampled from a set of known source domains $ d_1,d_2,...,d_n $, and the goal is to learn a meta-learning model that can generalize to an unseen target domain $ d_{n+1} $.
One of the main challenges in this setting is the low generalization ability of the meta-learning model to novel tasks sampled from target domains, i.e., the domain shift problem.

\subsection{Task-Augmented Meta-Learning}
Differences in environments can result in that different domains having different task distributions and holding different styles.
In CD-FSL, low generalization ability means that the model can not generalize well to new styles of target tasks from the unseen target domain. 
For example, the model, which is trained with source tasks sampled from MiniImageNet, tends to perform badly on CUB. That is due to the fact that, though the source domain (MiniImageNet) and the target domain (CUB) share some similar classes of birds, they differ in styles including resolution, color contrast, and illumination.

Based on the observation, we propose Task-Augmented Meta-Learning (TAML) to learn a more style-independent model, and the key idea is to design style transfer-based task augmentation and provide more styles of training tasks. Given a batch of tasks $ \mathbb{T}=\left\{T_j\right\}_{j=1}^n $, for the convenience in the following elaboration, we re-denote each task as $ T_j=(X_j,Y_j) $, where:
\begin{equation}
	\begin{aligned}
		X_j=(x_{1,j}^s,x_{2,j}^s,...,x_{N \times K_s,j}^s,x_{1,j}^q,x_{2,j}^q,...,x_{N \times K_q,j}^q ) ,
	\end{aligned}
\end{equation}
\begin{equation}
	\begin{aligned}
		Y_j=(y_{1,j}^s,y_{2,j}^s,...,y_{N \times K_s,j}^s,y_{1,j}^q,y_{2,j}^q,...,y_{N \times K_q,j}^q ).
	\end{aligned}
\end{equation}
And we denote $ X_j^l $ as features of task $ T_j $ output in $ l_{th} $ layer of the feature extractor.
Extra new tasks can be generated based on $ \mathbb{T} $ in the following form:
\begin{equation}
	\begin{aligned}
		\mathbb{T}^{new}=F(I(\mathbb{T}) \cup \mathbf{S}(\mathbb{T},I(\mathbb{T}))),
	\end{aligned}
\end{equation}
where $ I $ is the function for combining styles of original tasks and provide new styles for training, $ S $ refers to the function that transferring original tasks to new styles obtained by $ I $, and $ F $ imports uncertainty to each new tasks. Concretely in TAML, we propose Multi-Task Interpolation (MTI) as $ I $, Multi-Task Style Transfer (MTST) as $ \mathbf{S} $, and Feature Modulation (FM) as $ F $.
The diagram of our proposed TAML is shown in Figure \ref{fig:diagram}. MTI, MTST, and FM are proposed to perform style fusion, style transfer and random style transformation to achieve task augmentation correspondingly.
The proposed task augmentation is simple yet effective, and increases the diversity of training tasks to improve generalization. The details of the proposed task augmentation are introduced in Section \ref{MTI}, \ref{MTST} and \ref{FM} respectively.

\subsubsection{Task Augmentation by Multi-Task Interpolation}\label{MTI}
To improve the generalization ability of few-shot learning models, we propose MTI, a method that aims to generate more diverse styles of tasks for training.
Existing pair-wise techniques are limited to generating samples or tasks with similar styles.
Instead, MTI performs task interpolation on multiple original tasks, resulting in more diverse feature fusion and style fusion.
Combining the means and variances of multiple original tasks contributes to an interpolated task style that incorporates a combination of styles from the original tasks.
We focus on the most common non-label-sharing scenarios in few-shot learning, where classes are randomly assigned labels of $\left\{1,..,N\right\}$ in each task.
Therefore, interpolating the labels directly is meaningless.
In MTI, we conduct task interpolation at the feature level and re-assign new labels to the classes in each interpolated task.
Furthermore, when performing task interpolation, we interpolate both the query set and the support set together to ensure that they still share the same classes in each interpolated task, just like in original tasks.

In each iteration, given the batch of tasks $ \mathbb{T}=\left\{T_j\right\}_{j=1}^n $ sampled from source domains. Specifically, we randomly select $ m $ of these tasks to generate the interpolated task features:
\begin{equation}
	\begin{aligned}
		X^{mix,l}= \sum_{j=1}^m(\lambda_j X_j^l),
		\label{newtask1}
	\end{aligned}
\end{equation}
where $ m \in [2,n]$ and the weights $ \left\{\lambda_j\right\}_{j=1}^m $ are sampled from the predefined Dirichlet distribution:
\begin{equation}
	\begin{aligned}
		\mathbf{\lambda}=[\lambda_1,\lambda_2,...,\lambda_m] \sim Dirichlet(\mathbf{\gamma}).
	\end{aligned}
\end{equation}
The interpolated classes are regarded as totally new classes in the interpolated task, and with a re-assigned hot-vector of labels $ Y^{mix} $, we obtain a new task $ T^{mix,l}=(X^{mix,l},Y^{{mix}}) $. Since these interpolated classes may different from any classes in source domains, the interpolation can add number of classes and increase the diversity of tasks for training. More importantly, the style of $ T^{mix,l} $ will be the combination of the styles of $ \left\{T_j^l=(X_j^l,Y_j)\right\}_{j=1}^m $, which will be further utilized in the following section to learn a style-independent model. 

\begin{center}
	\begin{figure}[t]
		\centering
		\includegraphics[width=\linewidth]{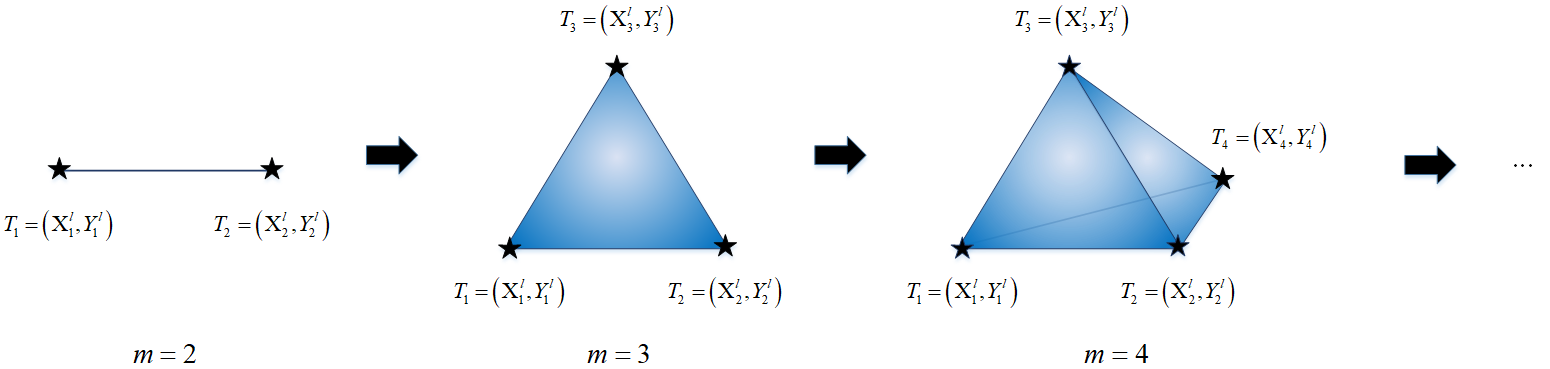}
		\caption{Illustration of Multi-Task Interpolation (MTI). When $ m=2 $, MTI only interpolates two tasks from the same domain, so interpolated tasks only exist on the edge of the triangle. When $ m \geq 3 $, MTI mixes tasks of multiple domains covering the whole area, meaning MTI introduces interpolated tasks with more information and higher diversity.}
		\label{fig:MTI}
	\end{figure}
\end{center}

The illustration of MTI is shown in Figure \ref{fig:MTI}. When $ m=2 $, MTI degenerates into MLTI \cite{yao2021meta}, which only interpolates pairs of tasks and generates tasks between two tasks (the lines between vertex). When $ m \geq 3 $, we can conduct task interpolation on more task, and interpolated tasks can be subject to a larger task distribution (the whole area). Given $ n $ tasks from source domains, pairwise task interpolation needs $ O(n \times n) $ interpolations to obtain all task combinations, and MTI needs only one interpolation. Thus, compared with recent work using Mixup or Manifold Mixup for interpolation \cite{ni2021data,yao2021improving,yao2021meta}, our method is more efficient and effective. 

\vspace{-1cm}
\subsubsection{Task Augmentation by Multi-Task Style Transfer}\label{MTST}
Building upon MTI, we propose MTST, which generates new tasks with different styles while preserving the original class information, thus enabling the learning of discriminative domain-independent or style-independent features.
By incorporating the original class information in source domains, MTST ensures that the model performs well on tasks both before and after style transfer, demonstrating its high domain generalization ability.

To achieve task style transfer, we need to obtain the style parameters of the task or domain based on the extracted features.
Previous studies have shown that high-level layers extract semantic information, while low-level layers process color and texture information \cite{gatys2015texture}.
Therefore, the low-level features of all images in a task can be considered the ``style" of the task or domain.
In MTST, we calculate the style parameters and perform style transfer using the low-level features.
This allows for effective style transfer while preserving the original class information.

\begin{center}
	\begin{figure}[t]
		\centering
		\includegraphics[width=.8\linewidth]{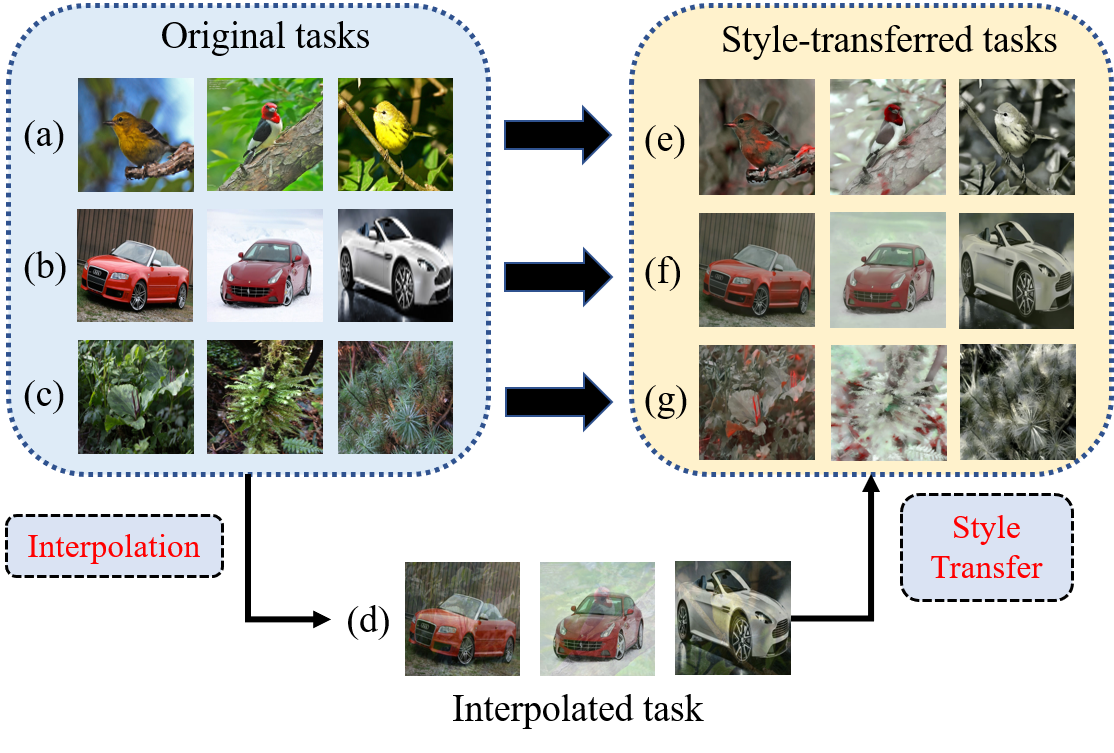}
		\caption{Visualization of new tasks. Take the interpolation and style transfer of three 3-way 1-shot tasks as an example. Based on original tasks (a-c), we conduct task interpolation to get task (d), the style of which is the fusion of original tasks. Then, we conduct style transfer on task (a-c), and get task (e-g) correspondingly.}
		\label{fig:new_tasks}
	\end{figure}
\end{center}

Before conducting style transfer, the style normalization can be achieved with the feature statistics of original task $ j $:
\begin{equation}
	\begin{aligned}
		\mu_j^{l} = \frac{1}{N}\sum_{i=1}^{N \times (K_s+K_q)}(x_{i,j}^{l}),
	\end{aligned}
\end{equation}
\begin{equation}
	\begin{aligned}
		\sigma_j^{l} = \frac{1}{N} \sum_{i=1}^{N \times (K_s+K_q)}(x_{i,j}^{l}-\mu_j^{l})^2.
	\end{aligned}
\end{equation}
The style parameters of the augmented domain can be obtained based on the interpolated task:
\begin{equation}
	\begin{aligned}
		\mu^{mix,l} = \frac{1}{N}\sum_{i=1}^{N \times (K_s+K_q)}(x_{i}^{mix,l}),
	\end{aligned}
\end{equation}
\begin{equation}
	\begin{aligned}
		\sigma^{mix,l} = \frac{1}{N} \sum_{i=1}^{N \times (K_s+K_q)}(x_{i}^{mix,l}-\mu^{mix,l})^2.
	\end{aligned}
\end{equation}
With $ \mu^{mix,l} $ and $ \sigma^{mix,l} $, style-transferred tasks are generated to augment the source domain by performing style transfer on the original task, 
\begin{equation}
	\begin{aligned}
		X^{transfer,l}_j= \dfrac{X_j^l-\mu_j^{l}}{\sigma_j^{l}} \cdot \sigma^{mix,l} + \mu^{mix,l}.
		\label{newtask2}
	\end{aligned}
\end{equation}
The style-transferred classes are also considered as entirely new classes in the style-transferred task.
By re-assigning hot-vector labels $Y^{transfer}_j$, we can generate a new style-transferred task $T^{transfer,l}_j = (X^{transfer,l}_j,Y^{{transfer}}_j)$, based on the interpolated task $T^{mix,l}$ and the original task $T_j$. We present the visualization of both original tasks and new tasks in Figure \ref{fig:new_tasks}.

Through task style transfer, the remapped task features can be better aligned with the characteristics of the augmented domain.
By mapping a task's features to the augmented domain, our model can learn to extract style-independent features and achieve better generalization performance.
Our method has several advantages:

1. Multi-Task Style transfer is a task-level domain augmentation method that increases the number of training tasks and enhances the generalization of the extracted features.

2. It enables the model to learn style-independent features by generating multiple cross-domain images with the same class information.

3. By using additional style-transferred images, our method can prevent model overfitting.

The style parameters are calculated with the interpolated task generated by MTI, which means that the whole performance of MTST depends on the performance of its important part MTI. Specifically, the parameters of Dirichelet distribution, i.e., $ \gamma $, which is used to provide the weights for task interpolation, will decide the style of new tasks.
Therefore, the selection of $ \gamma $ is essential.

\begin{center}
	\begin{figure}[t]
		\centering
		\includegraphics[width=\linewidth]{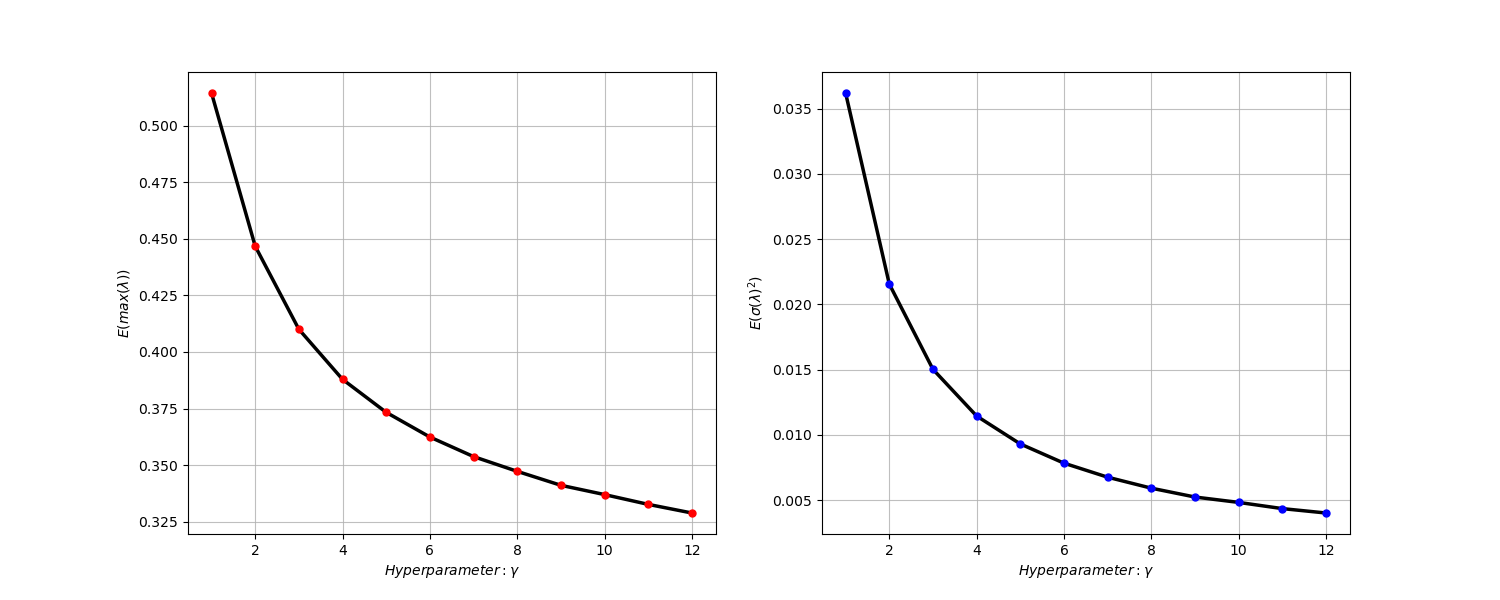}
		\caption{Study on the relationship between hyperparameter $ \gamma $ and the expectation of maximum and variance of $ \lambda $ when $ n=4 $. When $ \gamma $ gets larger, the generated tasks will become more random.}
		\label{fig:differ_gamma}
	\end{figure}
\end{center}

The relationship between $ \gamma $ and the expectation of maximum and variance of $ \lambda $ can be found in Figure \ref{fig:differ_gamma}, which shows that with the increase of $ \gamma $, generated samples become less definiteness and more random. The reason is that as $ \gamma $ increases, the Dirichelet distribution will approach a uniform distribution. If we want to focus more on domain $ j $ during interpolation, we can set $ \gamma_j $ larger than other components in $ \mathbf{\gamma} $, which assigns a larger weight $ \lambda_j $ to $ X_j^l $ statistically. When the target domain is very similar to one of source domains, the above parameters can be set. But in more cases, the similarity between the target domain and source domains is completely unknown. 
In the meta objective, the goal is to transfer knowledge from other domains and improve cross-domain generalization, which would be enhanced by interpolation results with larger domain discrepancy. So we can set $ \gamma_j $ smaller than other components in $ \mathbf{\gamma} $, which induces smaller $ \lambda_j $ statistically. In our experiments, we set the $ \gamma $ to be a vector of all $ 0.2 $ with dimension $ n $, since the target domain is totally unseen and we do not prefer a particular other domain.

\vspace{-2cm}

\subsubsection{Task Augmentation by Feature Modulation}\label{FM}
It is worth noting that, styles of new tasks $ X^{new,l}=\left\{X^{mix,l},X^{transfer,l}\right\} $, provided by MTST, can be regarded as linear combinations of styles of original source tasks. 
To introduce more random styles and simulate more feature distributions in the training stage, we propose FM to affline transform features of new tasks. 
Concretely, we introduce random parameters to modulate the features and improve the generalization of our model to the target domain. Firstly, we  sample the scaling and bias terms of affine transformations from Gaussian distributions,
\begin{equation}
	\begin{aligned}
		\alpha^l\sim N(0,softplus(W_{\alpha}^l)), l=1,...,L,
	\end{aligned}
\end{equation}
\begin{equation}
	\begin{aligned}
		\beta^{l}\sim N(0,softplus(W_{\beta}^l)), l=1,...,L,
	\end{aligned}
\end{equation}
where $ W_{\alpha}^l $ and  $W_{\alpha}^l $ denote learnable sampling hyper-parameters, and $ softplus(\cdot)=\log (1+exp(\cdot)) $ is the nonlinear activation function. We denote the parameters for FM as $  \theta_m=\left\{\mathbf{W_{\alpha}},\mathbf{W_{\beta}}\right\}$. We then compute the modulated features by applying the sampled affine transformations to intermediate features of layer $ X^{new,l} $ as follows:

\begin{equation}
	\begin{aligned}
		\hat{X}^{new,l} = X^{new,l}+\mathbf{\alpha}^l \cdot X^{new,l} + \mathbf{\beta}^l,
	\end{aligned}
\end{equation}
In practice, the same affine transformation is applied across all embeddings in the task $ T^{new,l} $. 
The feature-level data augmentation increases the diversity of training samples, thus dramatically reducing overfitting and improving stability and performance. FM is complementary to task-level augmentation and we apply both to help model training at the same time.

\subsubsection{The Algorithm}
We illustrate the process in Algorithm 1. The whole meta-training consists of two stages. In the first stage, we meta-train the model on the original tasks sampled from source domains.
In each epoch, given the batch of $ n $ tasks form source domains, initialization of feature extractor and classifier $ \theta_f,\theta_c $, we can get classification loss $ L^{SD} $ on query sets of these tasks:
\begin{equation}
	\begin{aligned}
		L^{SD} = \sum_{j=1}^{n}l(\theta_c(\theta_f(\mathcal{S}_j),\theta_f(\mathcal{Q}_j))),
	\end{aligned}
\end{equation}
where $ l $ is the standard cross-entropy in our experiments. Thus we can update feature extractor and classifier based on the meta loss, which means:
\begin{equation}
	\begin{aligned}
		(\theta'_f,\theta'_c)=(\theta_f,\theta_c)-lr \cdot \nabla_{\theta_f,\theta_c}L^{SD}.
		\label{updatefc}
	\end{aligned}
\end{equation}
where $ lr $ is the learning rate.

Then in the second stage, We improve the generalization ability of the updated network by meta-training it on the augmented source domains. By conducting MTST, we can get new tasks $ X^{new} $ on the augmented source domain, the detailed process of which has been illustrated in the above sections. After introducing FM $ \theta_m $, denote that the parameters for feature extracting as $ \theta'_{fm}=\left\{\theta'_f,\theta_m\right\}$. We can also calculate classification loss on query sets of these new tasks:
\begin{equation}
	\begin{aligned}
		L^{AD} = \sum_{j=1}^{n}l(\theta'_c(\theta'_{fm}(S^{new}_j),\theta'_{fm}(Q^{new}_j))).
	\end{aligned}
\end{equation}
And the model parameters can be updated on the augmented source domains, namely,
\begin{equation}
	\begin{aligned}
		(\theta''_f,\theta''_c, \theta'_m)=(\theta'_f,\theta'_c, \theta_m)-lr \cdot \nabla_{\theta'_f,\theta'_c,\theta_m}L^{AD}.
		\label{updatefcm}
	\end{aligned}
\end{equation}

\begin{algorithm}[t]
	\caption{TAML: online meta-training} 
	\label{alg:TAML}
	\begin{algorithmic}[1]
		\Require source domains: $d_1,...,d_n$, learning rate: $ lr $, hyper-parameters: $\gamma$;
		\Ensure {$\theta_f$,$\theta_c $, $ \theta_m $}\newline
		\textbf{Initialize} $\theta_f$,$\theta_c $, $ \theta_m $ ;
		\While{not none}
		\State Sample $ n $ tasks $ \left\{T_j\right\}_{j=1}^{n} $ from source domains;
		\State \textit{\textbf{Update meta-learner on source domains:}}
		\State Obtain $ L^{SD} $ by $ L^{SD} = \sum_{j=1}^{n}l(\theta_c(\theta_f(\mathcal{S}_j),\theta_f(\mathcal{Q}_j))) $;
		\State Obtain $ \theta'_f,\theta'_c $ by $ (\theta'_f,\theta'_c)=(\theta_f,\theta_c)-lr \cdot \nabla_{\theta_f,\theta_c}L^{SD} $;
		\State \textit{\textbf{Update meta-learner on augmented source domains:}}
		\State Generate interpolated tasks by:
		\State ~$X^{mix,l}= \sum_{j=1}^m(\lambda_j X_j^l)$;
		\State Calculate the style parameters of original tasks by: \State ~$\mu_j^{l} = \frac{1}{N}\sum_{i=1}^{N \times (K_s+K_q)}(x_{i,j}^{l})$,
		\State ~$\sigma_j^{l} = \frac{1}{N} \sum_{i=1}^{N \times (K_s+K_q)}(x_{i,j}^{l}-\mu_j^{l})^2$;
		\State Calculate the style parameters of interpolated tasks by: \State ~$\mu^{mix,l} = \frac{1}{N}\sum_{i=1}^{N \times (K_s+K_q)}(x_{i}^{mix,l})$,
		\State ~$\sigma^{mix,l} = \frac{1}{N} \sum_{i=1}^{N \times (K_s+K_q)}(x_{i}^{mix,l}-\mu^{mix,l})^2$;
		\State Generate style-transferred tasks by:
		\State ~$X^{transfer,l}_j= \dfrac{X_j^l-\mu_j^{l}}{\sigma_j^{l}} \cdot \sigma^{mix,l} + \mu^{mix,l}$;
		\State Obtain loss on the augmented source domains:
		\State ~$ L^{AD} = \sum_{j=1}^{n}l(\theta'_c(\theta'_{fm}(S^{new}_j),\theta'_{fm}(Q^{new}_j))) $;
		\State Update meta-learner:
		\State ~$(\theta''_f,\theta''_c, \theta'_m)=(\theta'_f,\theta'_c, \theta_m)-lr \cdot \nabla_{\theta'_f,\theta'_c,\theta_m}L^{AD}$ ;
		\EndWhile
	\end{algorithmic}
\end{algorithm}

\subsection{Theoretical Analysis}
We theoretically investigate how TAML improves the generalization performance with metric-based meta-learning methods. Specifically, we theoretically prove that TAML essentially induces a data-dependent regularizer on both categories of meta-learning methods and controls the Rademacher complexity \cite{yin2019rademacher}, leading to greater generalization. 
We also make a further comparison between $ m=2 $ and $ m\geq 3 $ in our MTI, and we provide the analysis in Appendix A.
For the simplicity of presentation, we analyze the generalization ability by considering the two-layer neural network with binary classification, and we denote $ X_j^{new,l} $ as $ X_j^{new} $, the parameters of the meta-learner as $ \theta $. 
The approximation of $ L^{AD} $ is obtained in the following lemma, which shows that $ L^{AD} $ is approximately $ L_{SD} $ plus regularization terms implicitly:

\textit{\textbf{Lemma.}} Consider the TAML with $ \lambda \sim Dirichlet(\gamma) $. For any  $ J \in \mathbb{N}_{+}  $, there exists a constant $ c>0 $, if $  c \mapsto d(y, c) $  is $  J-times $ differentiable for all y, the second order approximation of  $ L^{AD} $  is given by:
\begin{equation}
	\begin{aligned}
		L^{SD}+c \frac{1}{n} \sum_{j=1}^{n} \psi\left(X_{j}^{\top} \theta\right) \cdot 
		\theta^{\top}\left(\frac{1}{n} \sum_{j=1}^{n} X^{new}_{j} X_{j}^{new \top}\right) \theta 
	\end{aligned}
\end{equation}

\textbf{\textit{Proof. }} 
We have that the Taylor expansion of $ L_{AD} $ up to the second-order equals to:

\begin{equation}
	\begin{aligned}
		\frac{1}{n} \sum_{j=1}^{n} \mathcal{L}\left(\bar{\lambda} \mathcal{D}_{i}\right)+ c \frac{1}{n} \sum_{j=1}^{n} \psi\left(X_{j}^{\top} \theta\right) \theta^{\top} \operatorname{Cov}\left(X_{j}^{new} \mid X_{j}\right) \theta.
	\end{aligned}
\end{equation}
where:
\begin{equation}
	\begin{aligned}
		\frac{1}{n} \sum_{j=1}^{n} \mathcal{L}\left(\bar{\lambda} \mathcal{D}_{j}\right)
		=& \frac{1}{n} \mathcal{L}\left(\bar{\lambda}\left\{\mathcal{D}_{j}\right\}_{j=1}^{n}\right).
	\end{aligned}
\end{equation}	
Given the effect of batch normalization, the overall sample mean should be:
\begin{equation}
	\begin{split}
		\frac{1}{n} \sum_{j=1}^{n} X_{j}^{new}=0.
	\end{split}
\end{equation}
Thus, the covariance matrix can be obtained by:
\begin{equation}
	\begin{aligned}
		\operatorname{Cov}\left(X_{j}^{new} \mid X_{j}\right)=\frac{1}{n} \sum_{j=1}^{n} X^{new}_{j} X_{j}^{new \top}.
	\end{aligned}
\end{equation}	

According to the above Lemma, there exists an implicit regularization effect on $ \theta $, and we consider the regularization term in the following form:
\begin{equation}
	\begin{aligned}
		\mathcal{F}_R:=\left\{\mathbf{x} \mapsto \theta^{\top} \mathbf{x}: \mathbb{E}\left[\psi\left(\mathbf{x}^{\top} \theta\right)\right] \theta^{\top} \Sigma \theta \leq R\right\}.
		\label{reg}
	\end{aligned}
\end{equation}	
where $ \Sigma= \mathbb{E}\left[\mathbf{x}\mathbf{x}^{\top} \right] $.
Considering that:
\begin{equation}
	\begin{aligned}
		\theta^{\top}\Sigma\theta=Var(x)= [(\mu^{mix,l})^2+(\sigma^{mix,l})^2] (W_{\alpha}^l)^2,
		\label{var}
	\end{aligned}
\end{equation}		
$ R $ is directly related to statistics of new tasks, i.e., $ \mu^{mix,l} $ and $ \sigma^{mix,l} $.
Similarly as in \cite{zhang2020does}, the regularization term can be simplified as $ \|\theta\|_{\Sigma}^{2} \leq R $. The following theorem shows that this implicit regularization can reduce the Rademacher complexity \cite{bartlett2002rademacher} for better generalization:

\textbf{\textbf{Theorem.}} The generalization bound is:
\begin{equation}
	\mathcal{R}(\mathcal{F}_R) \leq \frac{\sqrt{R}\sqrt{rank(\Sigma)}}{\sqrt{n}}.
\end{equation}

\textbf{\textbf{Proof.}} Let $ \xi_{i} $ be independent uniform random variables taking values in $ \left\{-1,1\right\} $, i.e., Rademacher variables.
We can bound the empirical Rademacher complexity as follows:
\begin{equation}
	\begin{aligned}
		\hat{\mathcal{R}}_{n}\left(\mathcal{F}_{R}\right) 
		\leq \frac{\sqrt{R}}{n} \sqrt{\mathbb{E}_{X^{new}} \sum_{i=1}^{n}\left(X_{i}-X^{new}\right)^{\top} \Sigma^{\dagger}\left(X_{i}-X^{new}\right)}.
	\end{aligned}
	\label{proof1}
\end{equation}
Here $ \Sigma^{\dagger} $ denotes the Moore–Penrose inverse of $ \Sigma $. Using this bound on the empirical Rademacher complexity, we now bound the Rademacher complexity as follows:

\begin{equation}
	\begin{aligned}
		&\mathcal{R}_{n}\left(\mathcal{F}_{R}\right)=\mathbb{E}_{X_{1}, \ldots, X_{n}} \hat{\mathcal{R}}_{n}\left(\mathcal{F}_{R}\right) \\
		& \leq \mathbb{E}_{X_{1}, \ldots, X_{n}} \frac{\sqrt{R}}{n} \sqrt{\sum_{i=1}^{n} \mathbb{E}_{X^{new}}\left(X_{i}-X^{new}\right)^{\top} \Sigma^{\dagger}\left(X_{i}-X^{new}\right)} \\
		& \leq \frac{\sqrt{R}}{n} \sqrt{\sum_{i=1}^{n} \mathbb{E}_{X_{i}, X^{new}}\left(X_{i}-X^{new}\right)^{\top} \Sigma^{\dagger}\left(X_{i}-X^{new}\right)} \\
		&=\frac{\sqrt{R} \sqrt{\operatorname{rank}(\Sigma)}}{\sqrt{n}}.
	\end{aligned}
	\label{proof2}
\end{equation}

By using the law of total variance, we can know that:
\begin{equation}
	\begin{aligned}
		\operatorname{Cov}\left(X_{j}^{new} \mid X_{j}\right)
		&=  \mathbb{E}[\operatorname{Cov}\left(X_{j}^{new} \mid X_{j} \right)] +\operatorname{Cov}(\mathbb{E}[X_j^{new}|X_j]) \\
		&\geq \mathbb{E}[\operatorname{Cov}\left(X_{j}^{new} \mid X_{j} \right)]
	\end{aligned}
\end{equation}
The covariance matrix induced by style transfer-based task augmentation in TAML is $ \operatorname{Cov}(\mathbb{E}[X_j^{new}|X_j]) $. Thus in the proposed TAML, the regularization effect $ R $ will be smaller according to Eq. \ref{reg} and the generalization bound will be tighter, which means that the trained model can generalize better to unseen target domains. More detailed derivation can be seen in the appendix.

\section{Experiments}
In this section, details on the datasets employed along with implementation settings are presented. We evaluate our proposed method on two standard cross-domain few-shot benchmarks and compare results with recent state-of-the-art methods.
\subsection{Datasets}
The first benchmark is composed of five few-shot classification datasets from diverse domains: MiniImageNet (natural images, 100 classes), CUB \cite{wah2011caltech} (dataset of birds images, 200 classes), Cars \cite{krause20133d} (dataset of cars,196 classes), Places \cite{zhou2017places} (dataset of natural and human-made places, 365 classes), and Plantae \cite{van2018inaturalist} (dataset of plants, 200 classes).

The second benchmark is ECCV 2020 challenge, which also consists of five few-shot classification datasets : MiniImageNet, ChestX \cite{wang2017chestx} (dataset of X-ray images), ISIC \cite{codella2019skin} (dataset of dermoscopic images of skin lesions), EuroSAT \cite{helber2019eurosat} (dataset of satellite images), CropDisease \cite{mohanty2016using} (dataset of plant disease). 

Overall, the target datasets in the second benchmark are more challenging than the first benchmark.

\begin{table}[t]
	\centering
	\caption{Evaluative results ($ \% $) on the First Benchmark.}
	\label{tab:benchmark1}
	\renewcommand\arraystretch{1.25}
	\resizebox{\linewidth}{!}{
		\begin{threeparttable}
			\begin{tabular}{l|l|c|c|c|c}
				\hline
				\multicolumn{2}{c|}{1-shot}	&	CUB	&	Cars	&	Places	&	Plantae	\\
				\hline
				\multirow{3}{*}{MatchingNet\cite{vinyals2016matching}}		&FWT~\cite{tseng2020cross}				& 36.61$ \pm $0.53	& 29.82$ \pm $0.44	& 51.07$ \pm $0.68	& 34.48$ \pm $0.50	\\ 
				{}															&TAML(our)								& \textbf{40.37$ \pm $0.49}	& \textbf{33.62$ \pm $0.32}	& \textbf{54.09$ \pm $0.62}	& \textbf{37.42$ \pm $0.38}	 \\
				{}															&TAML\tnote{1} (our)					& \textbf{40.95$ \pm $0.47}	& \textbf{33.85$ \pm $0.32}	& \textbf{54.60$ \pm $0.57}	& \textbf{37.68$ \pm $0.34}	 \\\hline
				\multirow{4}{*}{RelationNet\cite{sung2018learning}}		&FWT~\cite{tseng2020cross}					& 44.07$ \pm $0.77	& 28.63$ \pm $0.59	& 50.68$ \pm $0.87	& 33.14$ \pm $0.62	\\
				{}														&ATA~\cite{wang2021cross}					& 43.02$ \pm $0.4	& 31.79$ \pm $0.3	& 51.16$ \pm $0.5	& 33.72$ \pm $0.3	\\ 
				{}														&TAML(our)			& \textbf{45.42$ \pm $0.58}	& \textbf{31.98$ \pm $0.46}	& \textbf{51.65$ \pm $0.62}	& \textbf{35.27$ \pm $0.53} \\
				{}														&TAML\tnote{1} (our)& \textbf{46.18$ \pm $0.52}	& \textbf{32.46$ \pm $0.43}	& \textbf{52.29$ \pm $0.55}	& \textbf{35.53$ \pm $0.48}	 \\\hline
				\multirow{6}{*}{GNN\cite{satorras2018few}}				&FWT~\cite{tseng2020cross}				& 47.47$ \pm $0.75	& 31.61$ \pm $0.53	& 55.77$ \pm $0.79	& 35.95$ \pm $0.58	\\ 
				{}									&ATA~\cite{wang2021cross}				& 45.00$ \pm $0.5	& 33.61$ \pm $0.4	& 53.57$ \pm $0.5	& 34.42$ \pm $0.4	\\ 
				{}									&LRP~\cite{sun2021explanation}				& 48.29$ \pm $0.51	& 32.78$ \pm $0.39	& 54.83$ \pm $0.56	& 37.49$ \pm $0.43	\\
				{}									&T3S~\cite{yuan2022task}				& 45.92	& 33.22	& 55.83	& -	\\
				{}									&TAML(our)			& \textbf{50.18$ \pm $0.76}	& \textbf{34.16$ \pm $0.38}	& \textbf{57.56$ \pm $0.64}	& \textbf{38.28$ \pm $0.47}	\\
				{}									&TAML\tnote{1} (our)	& \textbf{50.38$ \pm $0.64}	& \textbf{34.19$ \pm $0.36}	& \textbf{57.92$ \pm $0.52}	& \textbf{38.35$ \pm $0.46}	\\\hline\hline
				
				\multicolumn{2}{c|}{5-shot}	&	CUB	&	Cars	&	Places	&	Plantae	\\
				\hline
				\multirow{3}{*}{MatchingNet}		&FWT~\cite{tseng2020cross}				& 55.23$ \pm $0.83	& 41.24$ \pm $0.65	& 64.55$ \pm $0.75	& 41.69$ \pm $0.63	\\ 
				{}									&TAML(our)			& \textbf{57.04$ \pm $0.44}	& \textbf{43.84$ \pm $0.67}	& \textbf{65.89$ \pm $1.05}	& \textbf{43.17$ \pm $1.04	} \\
				{}									&TAML\tnote{1} (our)	& \textbf{57.67$ \pm $0.44}	& \textbf{44.43$ \pm $0.67}	& \textbf{66.38$ \pm $1.05}	& \textbf{43.89$ \pm $1.04}	 \\\hline
				\multirow{4}{*}{RelationNet}		&FWT~\cite{tseng2020cross}				& 59.46$ \pm $0.71	& 39.91$ \pm $0.69	& 66.28$ \pm $0.72	& 45.08$ \pm $0.59	\\ 
				{}									&ATA~\cite{wang2021cross}				& 59.36$ \pm $0.4	& 42.95$ \pm $0.4	& 66.90$ \pm $0.4	& 45.32$ \pm $0.3	\\ 
				{}									&TAML(our)								& \textbf{61.36$ \pm $0.44}	& \textbf{43.19$ \pm $0.67}	& \textbf{67.45$ \pm $1.05}	& \textbf{47.06$ \pm $1.04}	\\
				{}									&TAML\tnote{1} (our)					& \textbf{61.82$ \pm $0.44}	& \textbf{43.75$ \pm $0.67}	& \textbf{68.01$ \pm $1.05}	& \textbf{47.72$ \pm $1.04}	 \\\hline
				\multirow{6}{*}{GNN}				&FWT~\cite{tseng2020cross}				& 66.98$ \pm $0.68	& 44.90$ \pm $0.64	& 73.94$ \pm $0.67	& 53.85$ \pm $0.62	\\ 
				{}									&ATA~\cite{wang2021cross}				& 66.22$ \pm $0.5	& 49.14$ \pm $0.4	& 75.48$ \pm $0.4	& 52.69$ \pm $0.4	\\ 
				{}									&LRP~\cite{sun2021explanation}				& 64.44$ \pm $0.48	& 46.20$ \pm $0.46	& 74.45$ \pm $0.47	& 54.46$ \pm $0.46	\\
				{}									&T3S~\cite{yuan2022task}				& 69.16	& \textbf{49.82}	& 76.33	& -	\\
				{}									&TAML(our)						& \textbf{69.96$ \pm $0.51}	& 48.42$ \pm $0.49	& \textbf{76.49$ \pm $0.56}	& \textbf{56.85$ \pm $0.43}	\\
				{}									&TAML\tnote{1} (our)			& \textbf{70.54$ \pm $0.51}	& 49.01$ \pm $0.48	& \textbf{76.83$ \pm $0.56}	& \textbf{57.49$ \pm $0.43}	\\\hline
			\end{tabular}
			\begin{tablenotes}
				\item [1] We fine-tune the model on the target domain.
			\end{tablenotes}
		\end{threeparttable}
	}
\end{table}

\begin{figure*}[t]
	\centering
	\includegraphics[width=.85\linewidth]{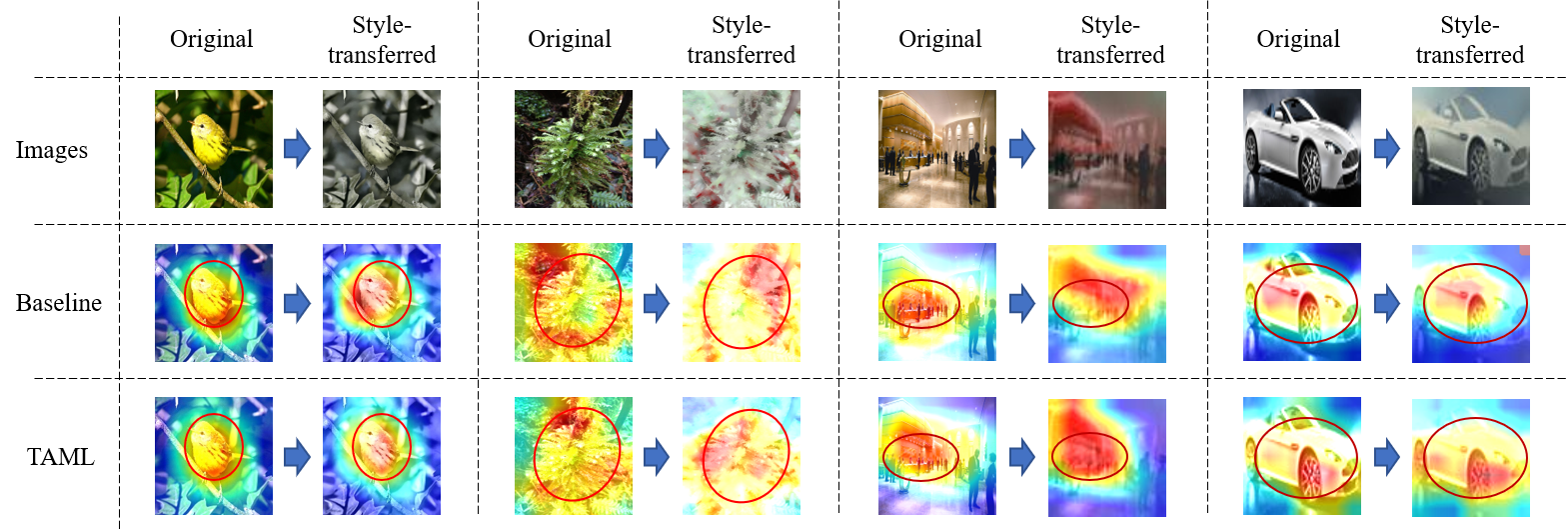}
	\caption{Activation maps of our TAML and the baseline w.r.t images of varied styles. The maps of TAML are more consistent and focus on the object for images of different styles.}
	\label{fig:style}
\end{figure*}

\subsection{Implementation details}
For a fair comparison, the ResNet-10 \cite{he2016deep} model is selected as the feature extractor in all experiments. 

The whole process contains three stages. In the first stage, the feature encoder is pre-trained by minimizing the cross-entropy classification loss on the 64 training classes in the MiniImageNet dataset.
The pre-training stage is the same as in \cite{tseng2020cross}. The second and third stages are the meta-training and meta-testing stages for few-shot classification, respectively. We conduct experiments on 5-way-1-shot and 5-way-5-shot settings. We use the Adam optimizer for training with the learning rate $ \alpha = 0.001 $.  In testing, 1000 episodes are randomly sampled from the target dataset to evaluate the model. The average classification accuracy and 95\% confidence interval are reported.

\subsection{Comparison to previous State-of-the-arts}
\subsubsection{Results on the First Benchmark}
To evaluate the effect on improving the cross-domain generalization ability on the first benchmark, three metric-based frameworks: MatchingNet \cite{vinyals2016matching}, RelationNet \cite{sung2018learning} and GNN \cite{satorras2018few} are taken into consideration for the classifier $ C $. We make detailed performance comparison of 5-way-1-shot and 5-way-5-shot tasks on the first benchmark with other methods, and provide the results in Table \ref{tab:benchmark1}.

\begin{table}[t]
	\centering
	\caption{The mean cosine similarity of tasks' features before and after style transfer features on the First Benchmark using GNN.}
	\label{tab:mean_cos}
	\renewcommand\arraystretch{1.25}
	\resizebox{.8\linewidth}{!}{
		\begin{tabular}{l|c|c|c|c}
			\hline
			Methods 	&	CUB		&	Cars	&	Places	&	Plantae	\\\hline
			Baseline 	&	0.6833233	&	0.6473708	&0.72646636 & 0.5987049\\\hline
			TAML	&0.7273805	&	0.7018726 &  0.73469505	& 0.6848321 \\
			\hline
		\end{tabular}
	}
\end{table}
\begin{table}[t]
	\centering
	\caption{Evaluative results ($ \% $) on the First Benchmark in the leave-one-out setting.}
	\label{tab:benchmark1_multiple}
	\renewcommand\arraystretch{1.25}
	\resizebox{\linewidth}{!}{
		\begin{threeparttable}
			\begin{tabular}{l|l|c|c|c|c}
				\hline
				\multicolumn{2}{c|}{1-shot}	&	CUB	&	Cars	&	Places	&	Plantae	\\
				\hline
				\multirow{3}{*}{MatchingNet\cite{vinyals2016matching}}		&FWT~\cite{tseng2020cross}	& 43.29$ \pm $0.59	& 30.62$ \pm $0.48	& 52.51 $ \pm $ 0.67	& 35.12 $ \pm $ 0.54	\\ 
				{}															&TAML(our)					& \textbf{45.07$ \pm $0.50}	& \textbf{31.46$ \pm $0.45}	& \textbf{54.46 $ \pm $ 0.52}	& \textbf{36.30$ \pm $ 0.48}	 \\
				{}															&TAML\tnote{1} (our)		& \textbf{45.82$ \pm $0.48}	& \textbf{32.34$ \pm $0.42}	& \textbf{54.90 $ \pm $ 0.49}	& \textbf{37.12$ \pm $ 0.46}	 \\\hline
				\multirow{3}{*}{RelationNet\cite{sung2018learning}}			&FWT~\cite{tseng2020cross}	& 48.38$ \pm $0.63	& 32.21$ \pm $0.51	& 50.74 $ \pm $ 0.66	& 35.00 $ \pm $ 0.52	\\
				{}															&TAML(our)					& \textbf{49.93$ \pm $0.55}	& \textbf{32.89$ \pm $0.44}	& \textbf{52.34 $ \pm $ 0.47}	& \textbf{36.27$ \pm $ 0.46} \\
				{}															&TAML\tnote{1} (our)		& \textbf{50.57$ \pm $0.52}	& \textbf{33.65$ \pm $0.42}	& \textbf{52.73 $ \pm $ 0.53}	& \textbf{36.95$ \pm $ 0.40}	 \\\hline
				\multirow{4}{*}{GNN\cite{satorras2018few}}					&FWT~\cite{tseng2020cross}	& 51.51$ \pm $0.80	& 34.12 $ \pm $ 0.63	& 56.31 $ \pm $ 0.80	& 42.09 $ \pm $ 0.68	\\ 
				{}															&LR2Net~\cite{chen2022cross-domain}	& 52.04$ \pm $0.70	& 34.84 $ \pm $ 0.62	& 57.57 $ \pm $ 0.78	& 42.05 $ \pm $ 0.70	\\ 
				{}															&TAML(our)					& \textbf{53.00$ \pm $0.74}	& \textbf{34.62 $ \pm $ 0.59}	& \textbf{56.80 $ \pm $ 0.68}	& \textbf{43.25 $ \pm $ 0.62}	\\
				{}															&TAML\tnote{1} (our)		& \textbf{53.62$ \pm $0.74}	& \textbf{35.37 $ \pm $ 0.50}	& \textbf{57.16$ \pm $ 0.57}	& \textbf{43.89 $ \pm $ 0.61}	\\\hline\hline
				
				\multicolumn{2}{c|}{5-shot}	&	CUB	&	Cars	&	Places	&	Plantae	\\
				\hline
				\multirow{3}{*}{MatchingNet}		&FWT~\cite{tseng2020cross}	& 61.41 $ \pm $ 0.57	& 43.08 $ \pm $ 0.55	& 64.99 $ \pm $ 0.59	& 48.32 $ \pm $ 0.57	\\ 
				{}									&TAML(our)					& \textbf{63.08$ \pm $ 0.52}	&\textbf{43.73 $ \pm $ 0.48	}&\textbf{66.48 $ \pm $ 0.53}	&\textbf{49.09 $ \pm $ 0.50} \\
				{}									&TAML\tnote{1} (our)		& \textbf{63.59$ \pm $ 0.50}	&\textbf{44.58 $ \pm $ 0.43}	&\textbf{66.87 $ \pm $ 0.47}	&\textbf{49.70 $ \pm $ 0.44}	 \\\hline
				\multirow{3}{*}{RelationNet}		&FWT~\cite{tseng2020cross}	& 64.99 $ \pm $ 0.54	&43.44 $ \pm $ 0.59	&67.35 $ \pm $ 0.54	&50.39 $ \pm $ 0.52	\\ 
				{}									&TAML(our)					& \textbf{66.52 $ \pm $ 0.45}	&\textbf{43.95 $ \pm $ 0.50}	&\textbf{68.66 $ \pm $ 0.49}	&\textbf{50.99 $ \pm $ 0.49}\\
				{}									&TAML\tnote{1} (our)		& \textbf{67.14$ \pm $ 0.45}	&\textbf{44.88 $ \pm $ 0.47}	&\textbf{69.15 $ \pm $ 0.48}	&\textbf{51.46 $ \pm $ 0.42}	 \\\hline
				\multirow{4}{*}{GNN}				&FWT~\cite{tseng2020cross}	& 73.11 $ \pm $ 0.68	&49.88 $ \pm $ 0.67	&77.05 $ \pm $ 0.65	&58.84 $ \pm $ 0.66	\\ 
				{}									&LR2Net~\cite{chen2022cross-domain}					& 73.94 $ \pm $ 0.68	&50.63 $ \pm $ 0.70	&76.68 $ \pm $ 0.61	&62.14 $ \pm $ 0.69	\\ 
				{}									&TAML(our)					& \textbf{74.46 $ \pm $ 0.49}	&\textbf{50.25 $ \pm $ 0.62}	&\textbf{77.35 $ \pm $ 0.56}	&\textbf{59.72 $ \pm $ 0.59}	\\
				{}									&TAML\tnote{1} (our)		&\textbf{74.90 $ \pm $ 0.48}	&\textbf{50.86 $ \pm $ 0.59}	&\textbf{77.69 $ \pm $ 0.54}	&\textbf{60.16 $ \pm $ 0.48}	\\\hline
			\end{tabular}
			\begin{tablenotes}
				\item [1] We fine-tune the model on the target domain.
			\end{tablenotes}
		\end{threeparttable}
	}
\end{table}

From the results, we have the following observations. Our TAML outperforms all the latest methods in most cases. 
The competitors, including FWT, LRP, ATA, are all specifically designed for CD-FSL. When comparing our TAML to these CD-FSL competitors, our method shows obvious advantages in most cases. 
The superior performance on target datasets of our methods demonstrates that the proposed style transfer-based task augmentation is effective in
reducing the domain gap between the source and target datasets in CD-FSL. 
By expanding the diversity of styles of source tasks, our method can address style shift between source and target tasks, and improve the generalization ability of the model.

To verify that we can extract more style-independent features than the strong baseline (FWT), we compare the mean cosine similarity of original tasks' features and style-transferred tasks' features in Table \ref{tab:mean_cos}. Results show that, features of original tasks and style-transferred tasks extracting by TAML are more similar, especially on Cars and Plantae. Moreover, in Figure \ref{fig:style}, we further compare the activation maps  \cite{zhou2016learning} of different methods by varying styles of input images. We can see that, for the images with different styles, the activation maps of TAML are more consistent than those of the baseline. The activation maps of baseline are more disorganized and are easily affected by style variants. These indicate that the model trained by TAML is more robust to style variations.

\begin{figure}[h]
	\centering
	\subfigure[Cub]{
		\includegraphics[width=.43\linewidth]{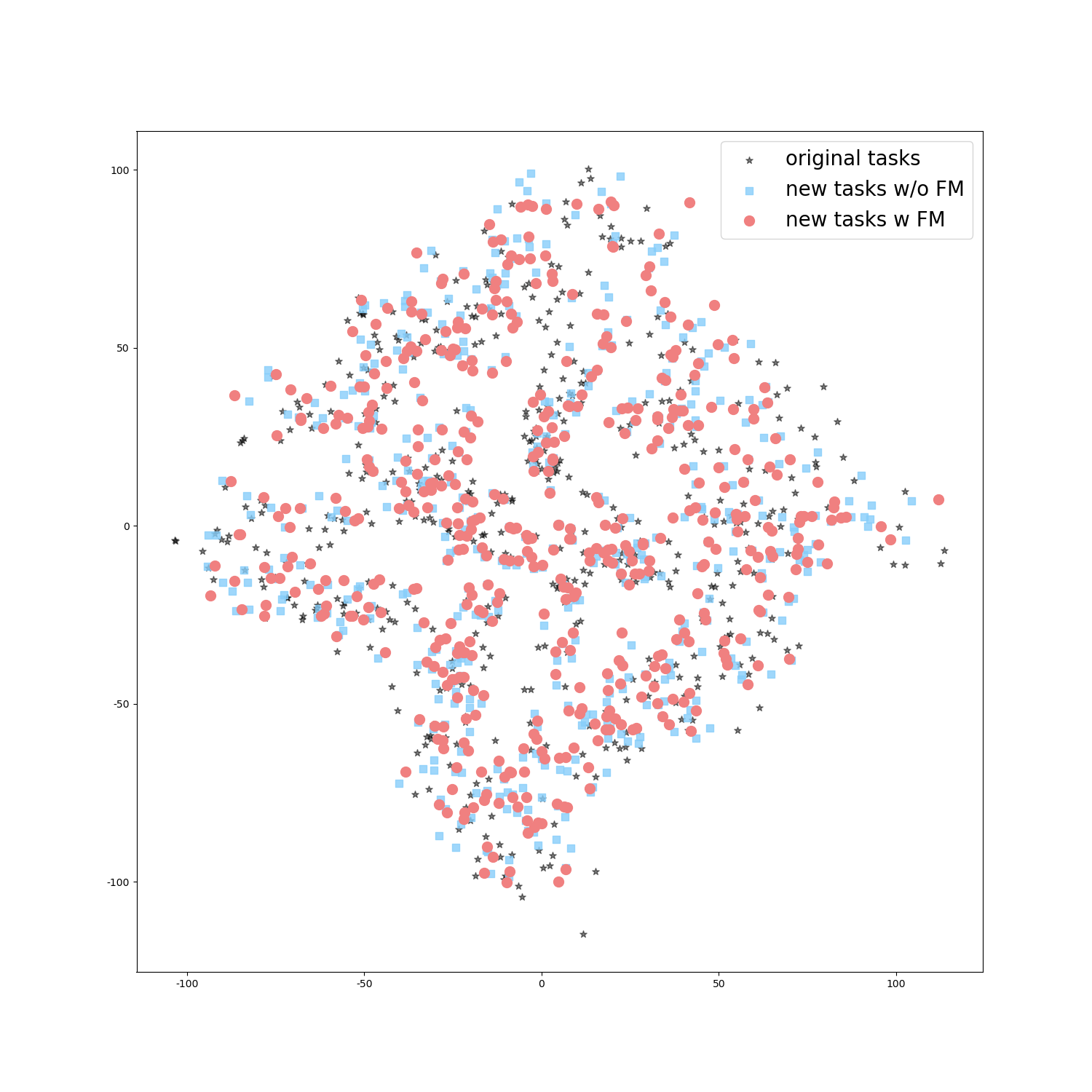}
	}
	\subfigure[Cars]{
		\includegraphics[width=.43\linewidth]{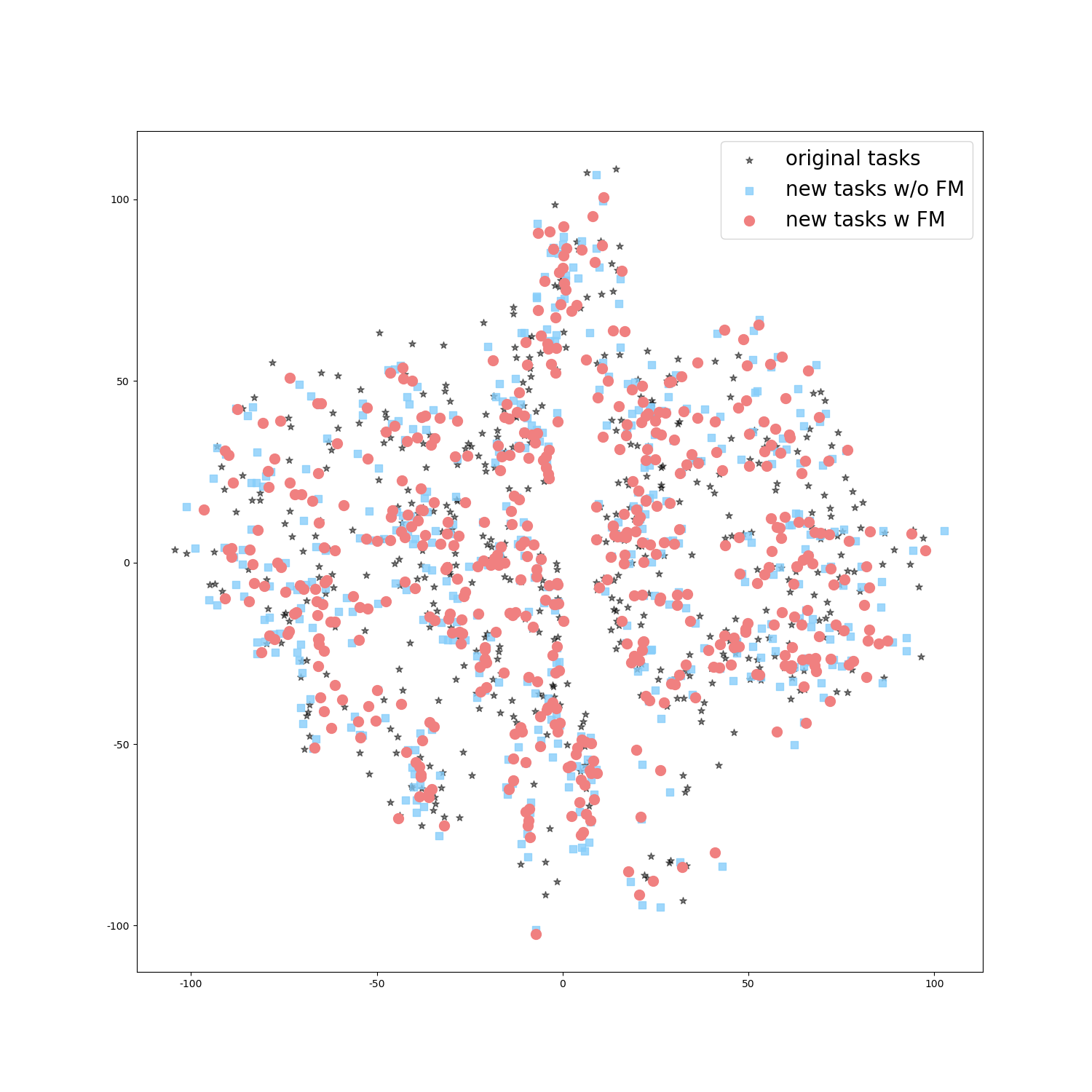}
	}
	\subfigure[Places]{
		\includegraphics[width=.43\linewidth]{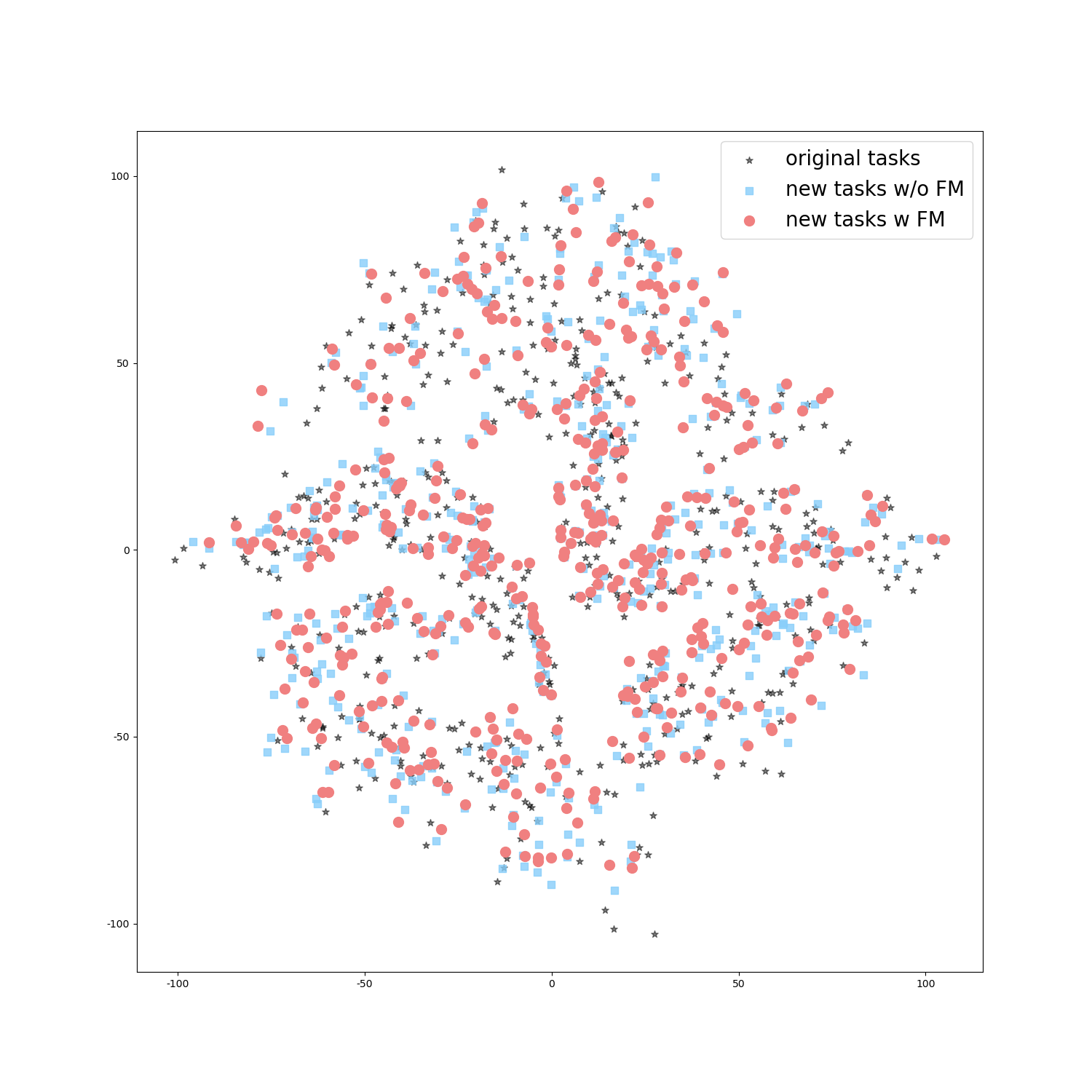}
	}
	\subfigure[Plantae]{
		\includegraphics[width=.43\linewidth]{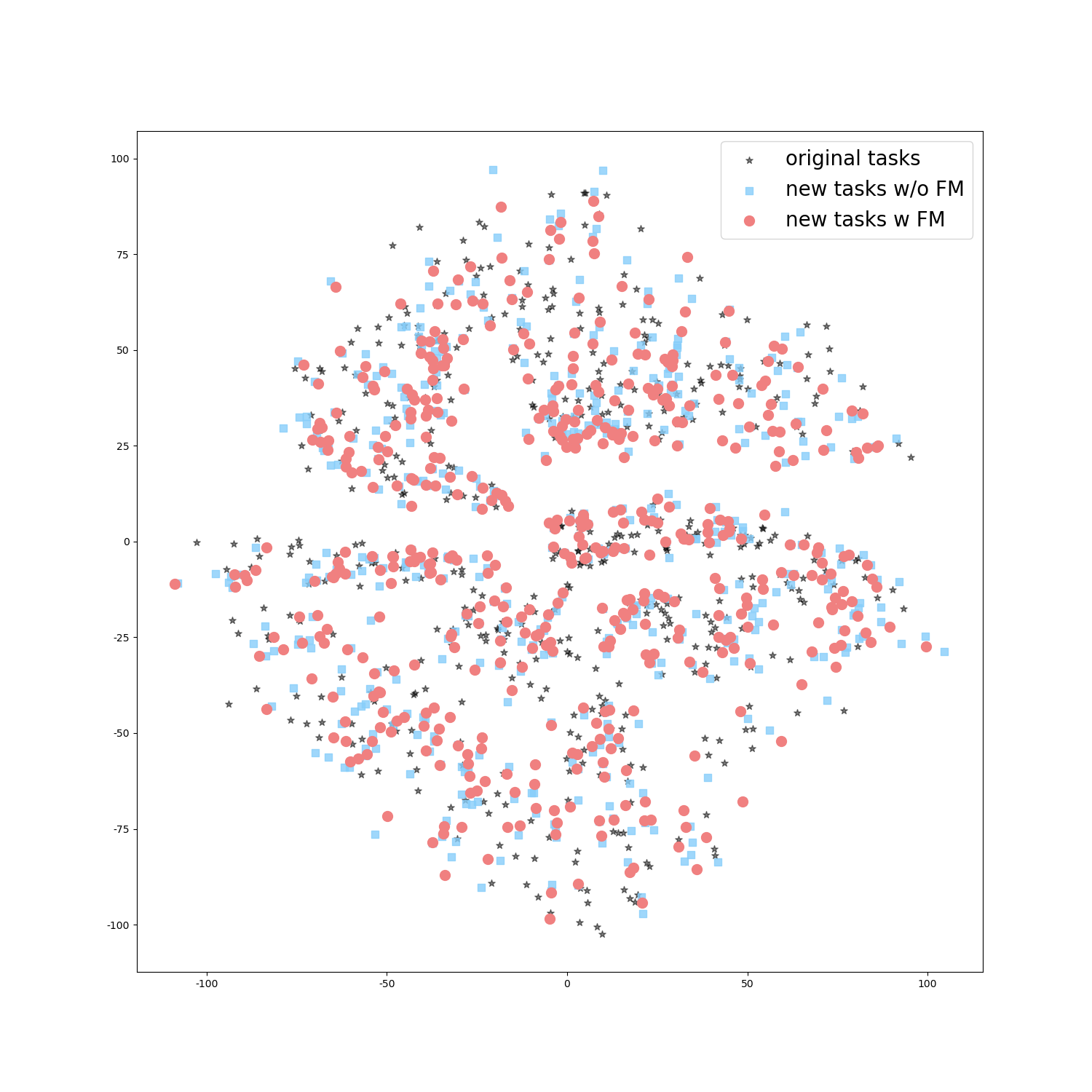}
	}
	\caption{Visualization of training tasks on the first benchmark using the leave-one-out strategy. New tasks generated by conducting interpolation and style transfer on original tasks densify the distribution of training tasks effectively.}
	\label{fig:tasks}
\end{figure}

We also conduct training on multiple source domains using the leave-one-out strategy in \cite{tseng2020cross}, and results are reported in Table \ref{tab:benchmark1_multiple}. TAML also achieves excellent performance compared with other methods in this setting, which demonstrates that our methods can ultilize information from various source domains and generalize well to different target domain.
We show the visualization of representations of both original tasks and interpolated tasks in Figure \ref{fig:tasks}. Specifically, we randomly select 1000 original tasks and new tasks generated by them under the 1-shot setting.
Each task is represented by the average of its prototypes. The figure suggests that the interpolated tasks generated by TAML indeed densify the task distribution and bridge the gap between different tasks.

\begin{table}[t]
	\centering
	\caption{Evaluative results ($ \% $) on the Second Benchmark.}
	\label{tab:benchmark2}
	\renewcommand\arraystretch{1.25}
	\resizebox{\linewidth}{!}{		
	\begin{threeparttable}
		\begin{tabular}{l|c|c|c|c}
			\hline
			1-shot &	ChestX	&	ISIC	&	EuroSAT	&	CropDisease	\\
			\hline
			FWT\cite{tseng2020cross}			& 22.04$ \pm $0.44	& 31.58$ \pm $0.67	& 62.36$ \pm $1.05	& 66.36$ \pm $1.04	\\ 
			ATA\cite{wang2021cross}			& 22.10$ \pm $0.20	& 33.21$ \pm $0.40	& 61.35$ \pm $0.50	& 67.47$ \pm $0.50	\\ 
			Meta-FDMixup\cite{fu2021meta}& 22.26$ \pm $0.45	& 32.48$ \pm $0.64	& 62.97$ \pm $1.01	& 66.23$ \pm $1.03	\\
			LRP\cite{sun2021explanation}			& 22.11$ \pm $0.20	& 30.94$ \pm $0.30	& 54.99$ \pm $0.50	& 59.23$ \pm $0.50	\\
			AFA	\cite{hu2022adversarial}			& \textbf{22.92$ \pm $0.2}	& 33.21$ \pm $0.3	& 63.12$ \pm $0.5	& 67.61$ \pm $0.5	\\ \hline
			TAML (ours) & 22.55$ \pm $0.36	& \textbf{33.30$ \pm $0.42}	& \textbf{65.24$ \pm $0.75}	& \textbf{69.48$ \pm $0.74}	\\
			TAML\tnote{1} (ours) & 22.67$ \pm $0.40	& \textbf{33.35$ \pm $0.39}	& \textbf{65.59$ \pm $0.76}	& \textbf{69.72$ \pm $0.80}	\\
			\hline
			\hline
			5-shot &	ChestX	&	ISIC	&	EuroSAT	&	CropDisease	\\
			\hline
			
			FWT\cite{tseng2020cross}			& 25.18$ \pm $0.45	& 43.17$ \pm $0.70	& 83.01$ \pm $0.79	& 87.11$ \pm $0.67	\\ 
			ATA\cite{wang2021cross}			& 24.32$ \pm $0.40	& 44.91$ \pm $0.40	& 83.75$ \pm $0.40	& 90.59$ \pm $0.30	\\ 
			Meta-FDMixup\cite{fu2021meta}& 24.52$ \pm $0.44	& 44.28$ \pm $0.66	& 80.48$ \pm $0.79	& 87.27$ \pm $0.69	\\
			LRP\cite{sun2021explanation}			& 24.53$ \pm $0.30	& 44.14$ \pm $0.40	& 77.14$ \pm $0.40	& 86.15$ \pm $0.40	\\
			AFA	\cite{hu2022adversarial}			& 25.02$ \pm $0.2	& 46.01$ \pm $0.4	& \textbf{85.58$ \pm $0.4}	& 88.06$ \pm $0.3	\\ \hline
			TAML (ours) & \textbf{25.87$ \pm $0.40}	& \textbf{46.16$ \pm $0.46}	& 85.09$ \pm $0.82	& \textbf{89.32$ \pm $0.90}\\
			TAML\tnote{1} (ours) & \textbf{26.15$ \pm $0.42}	& \textbf{46.68$ \pm $0.48}	& 85.52$ \pm $0.80	& \textbf{89.74$ \pm $0.93}	\\
			\hline
		\end{tabular}
		\begin{tablenotes}
			\item [1] We fine-tune the model on the target domain.
		\end{tablenotes}
	\end{threeparttable}
	}
\end{table}

\vspace{-1cm}
\subsubsection{Results on the Second Benchmark}
We further compare our method against latest methods on the second benchmark. Results are reported in Table \ref{tab:benchmark2}.
When the domain gap is getting relative larger, the performance gain of previous CD-FSL methods (i.e., FWT, ATA, Meta-FDmixup, LRP, AFA) are more limited. 
In contrast, our TAML achieves superior performance, which indicates that addressing the style shift between source tasks and target tasks can make a great contribution to improving the generalization to target domains.

\begin{table}[t]
	\centering
	\caption{Ablation Study ($ \% $) on the First Benchmark.}
	\label{tab:ablation_benchmark1}
	\renewcommand\arraystretch{1.25}
	\resizebox{\linewidth}{!}{
		\begin{tabular}{l|c|c|c|c|c}
			\hline
			\multicolumn{2}{c|}{1-shot}	&	CUB	&	Cars	&	Places	&	Plantae	\\
			\hline
			\multirow{7}{*}{MatchingNet\cite{vinyals2016matching}}		&Baseline 		& 36.61$ \pm $0.54  & 29.82$ \pm $0.52	& 52.47$ \pm $0.63	& 34.92$ \pm $0.49\\\cline{2-6}
			{}															&TI							& 38.80$ \pm $0.52	& 31.96$ \pm $0.44	& 53.00$ \pm $0.68	& 36.10$ \pm $0.50	\\ 
			{}															&TST						& 39.20$ \pm $0.55	& 32.62$ \pm $0.32	& 52.98$ \pm $0.57	& 36.45$ \pm $0.34\\
			{}															&TI+TST						& 39.83$ \pm $0.53	& 32.92$ \pm $0.32	& 53.67$ \pm $0.57	& 36.86$ \pm $0.34\\\cline{2-6}
			{}															&MTI						& 39.42$ \pm $0.53	& 32.74$ \pm $0.32	& 53.93$ \pm $0.62	& 36.75$ \pm $0.36	 \\ 
			{}															&MTST						& 39.66$ \pm $0.56	& 33.18$ \pm $0.32	& 53.64$ \pm $0.57	& 37.03$ \pm $0.39\\
			{}															&MTI + MTST	& \textbf{40.37$ \pm $0.49}	& \textbf{33.62$ \pm $0.32}	& \textbf{54.09$ \pm $0.62}	& \textbf{37.42$ \pm $0.38	} \\\hline
			\multirow{7}{*}{RelationNet\cite{sung2018learning}}			&Baseline		& 43.28$ \pm $0.68  & 28.72$ \pm $0.51	& 49.85$ \pm $0.69	& 32.79$ \pm $0.58\\\cline{2-6}
			{}															&TI							& 44.15$ \pm $0.64	& 29.86$ \pm $0.47	& 50.35$ \pm $0.62	& 33.54$ \pm $0.54	\\ 
			{}															&TST						& 44.30$ \pm $0.68	& 30.04$ \pm $0.50	& 50.46$ \pm $0.68	& 33.75$ \pm $0.57\\
			{}															&TI+TST						& 44.97$ \pm $0.66	& 30.63$ \pm $0.51	& 51.23$ \pm $0.65	& 34.49$ \pm $0.57\\\cline{2-6}
			{}															&MTI						& 44.57$ \pm $0.62	& 30.19$ \pm $0.46	& 50.88$ \pm $0.62	& 34.06$ \pm $0.46	 \\
			{}															&MTST						& 44.81$ \pm $0.67	& 30.52$ \pm $0.52	& 50.96$ \pm $0.67	& 34.26$ \pm $0.54\\
			{}															&MTI + MTST			& \textbf{45.42$ \pm $0.58}	& \textbf{31.98$ \pm $0.46}	& \textbf{51.65$ \pm $0.62}	& \textbf{35.27$ \pm $0.53	} \\\hline
			\multirow{7}{*}{GNN\cite{satorras2018few}}					&Baseline		& 47.02$ \pm $0.70  & 31.60$ \pm $0.49	& 55.22$ \pm $0.72	& 35.93$ \pm $0.55\\\cline{2-6}
			{}															&TI							& 48.68$ \pm $0.64	& 32.68$ \pm $0.44	& 56.14$ \pm $0.68	& 37.23$ \pm $0.50	\\ 
			{}															&TST						& 48.82$ \pm $0.68	& 33.25$ \pm $0.47	& 55.96$ \pm $0.73	& 37.26$ \pm $0.54\\
			{}															&TI+TST						& 49.64$ \pm $0.64	& 33.69$ \pm $0.46	& 56.81$ \pm $0.69	& 37.79$ \pm $0.53\\\cline{2-6}
			{}															&MTI						& 49.46$ \pm $0.63	& 33.75$ \pm $0.36	& 56.55$ \pm $0.64	& 37.50$ \pm $0.46	 \\
			{}															&MTST						& 49.58$ \pm $0.72	& 33.79$ \pm $0.44	& 56.68$ \pm $0.70	& 37.73$ \pm $0.51\\
			{}															&MTI + MTST		& \textbf{50.18$ \pm $0.76}	& \textbf{34.16$ \pm $0.38}	& \textbf{57.56$ \pm $0.64}	& \textbf{38.28$ \pm $0.47	} \\\hline
		\end{tabular}
	}
\end{table}

\begin{table}[t]
	\centering
	\caption{Ablation Study ($ \% $) on the Second Benchmark.}
	\label{tab:abalation_benchmark2}
	\renewcommand\arraystretch{1.25}
	
	\resizebox{\linewidth}{!}{
		\begin{tabular}{c|c|c|c|c}
			\hline
			1-shot &	ChestX	&	ISIC	&	EuroSAT	&	CropDisease	\\
			\hline
			FT-baseline		& 22.01$ \pm $0.49	& 31.32$ \pm $0.56	& 61.02$ \pm $0.89	& 65.90$ \pm $0.78	\\ \cline{1-5}
			TI				& 22.20$ \pm $0.42	& 32.18$ \pm $0.50	& 62.64$ \pm $0.69	& 67.07$ \pm $0.72	\\ 
			TST				& 22.32$ \pm $0.47	& 32.27$ \pm $0.52	& 62.86$ \pm $0.74	& 67.29$ \pm $0.77	\\
			TI+TST			& 22.43$ \pm $0.47	& 32.61$ \pm $0.53	& 63.75$ \pm $0.73	& 67.97$ \pm $0.77	\\\cline{1-5}
			MTI				& 22.39$ \pm $0.38	& 32.55$ \pm $0.46	& 63.26$ \pm $0.65	& 67.63$ \pm $0.68	\\ 
			MTST			& 22.44$ \pm $0.39	& 32.62$ \pm $0.46	& 63.52$ \pm $0.78	& 67.85$ \pm $0.75	\\ \cline{1-5}
			MTI + MTST & \textbf{22.55$ \pm $0.36}	& \textbf{33.30$ \pm $0.42}	& \textbf{65.24$ \pm $0.75}	& \textbf{69.48$ \pm $0.74}	\\
			\hline
		\end{tabular}
	}
\end{table}

\subsection{Ablation Study}
\subsubsection{Verification of Each Component}
We conduct experiments on the first benchmark to show the effectiveness of each proposed component, especially the necessity of introducing the Dirichlet distribution to interpolating multiple tasks and transferring the style of original tasks.

In Table \ref{tab:ablation_benchmark1} and \ref{tab:abalation_benchmark2}, we compare the performance of pair-wise Task Interpolation (TI) and Task Style Transfer (TST) with our MTI and MTST. Results on different target datasets show that our methods perform better than mixup-based methods, which infers that integration of original tasks on the source domains can effectively improve the generalization of the model. It is worth noting that style-based methods (i.e., TST and MTST) are checked to be more effective than interpolation-based methods (i.e., TI and MTI). The reason is that in the cross-domain few-shot learning, the domain shift mainly exists in the style variation of different datasets.
\begin{center}
	\begin{figure*}[t]
		\centering
		\includegraphics[width=.6\linewidth]{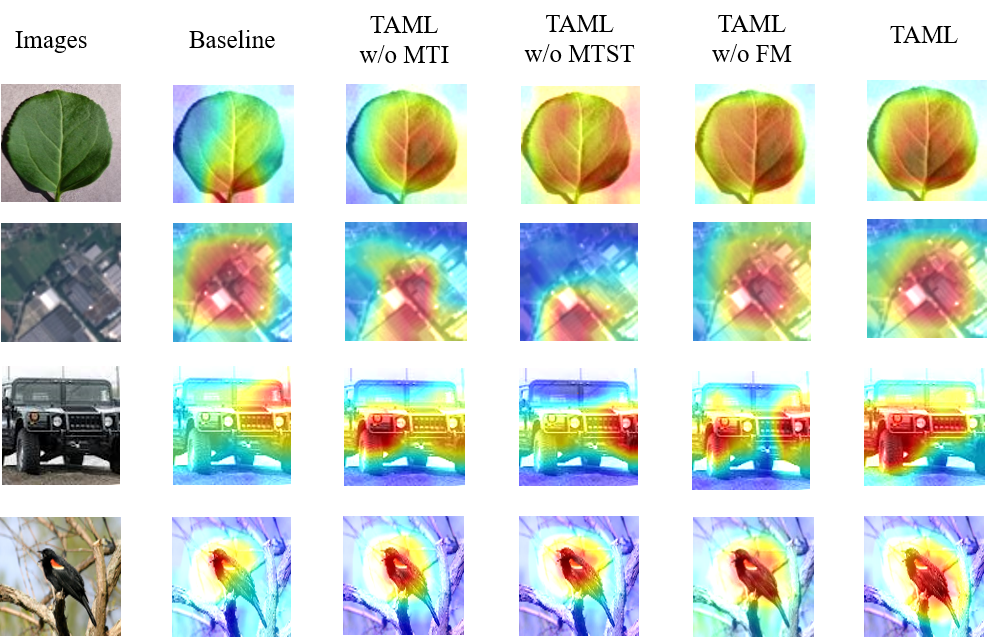}
		\caption{Class activation maps of different methods for CUB, Cars, CropDisease, EuroSAT 1-shot. The maps of TAML are more focus on the object for images from different domains.}
		\label{fig:cam}
	\end{figure*}
\end{center}
\begin{center}
	\begin{figure*}[h]
		\centering
		\includegraphics[width=\linewidth]{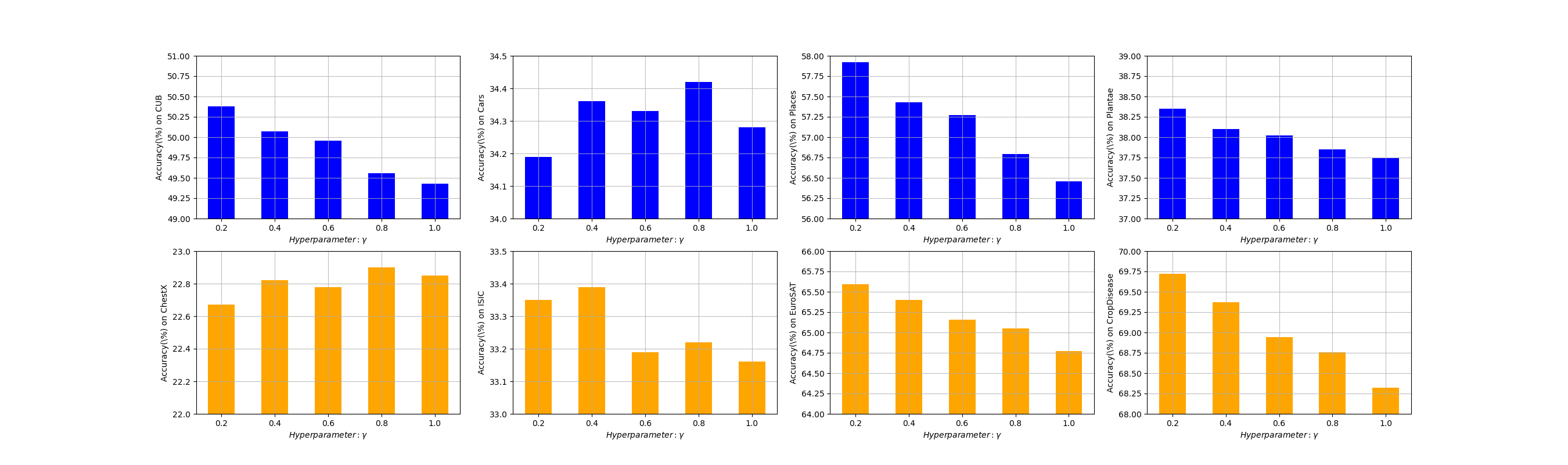}
		\caption{Hyperparameter Analysis on the $ \gamma $ of the Dirichlet distribution for different target domain. When $ \gamma $ changes from 0 to 1, the average accuracy will drop, which may be caused by too much noise in new tasks. }
		\label{fig:differ_gamma_ablation}
	\end{figure*}
\end{center}
\vspace{-1cm}
We utilize the class activation maps to visualize features noted in the classification process. We randomly choose CUB, Cars, CropDisease, EuroSAT from two benchmarks, and results are reported in Figure \ref{fig:cam}. Results shows that the baseline tends to be affected by environmental factors, and the extracted feature region is not complete enough sometimes. While our proposed methods obtain activation on more accurate discriminative regions, and contribute to focusing on some new feature regions beyond the baseline.


\subsubsection{Hyperparameter Analysis}
We also conduct experiments to study the impact of hyper-parameter $ \gamma $ which influence the strength of MTI in Eq.(8).
The results are shown in Figure  \ref{fig:differ_gamma_ablation}. We can find that there exists a drop in average accuracy when $ \gamma $ changes from 0 to 1. Such result is consistent to our argument that when we choose a large $ \gamma $ value, the expectation of its maximum and variance tend to be small, which may lead to too much noise.

\section{Conclusions}
In this work, to bridge the domain shift between the source domains and the target domain, we propose Task Augmented Meta-Learning (TAML) to conduct style transfer-based task augmentation in source domains.
Concretely, we introduce Multi-Task Interpolation (MTI) to interpolate multiple tasks and provide more task styles. Based on MTI, the Multi-Task Style Transfer (MTST) is proposed to transfer original tasks to new styles, which contributes to learning style-independent features. In addition, the proposed Feature Modulation (FM) imports uncertainty to features and provides more styles of new tasks for training.
Theoretical analysis shows that our method can definitely tighten the generalization bound and improve the generalization ability of the model.
We conduct extensive experiments on two cross-domain few-shot benchmarks, and our TAML achieves new state-of-the-art results on most datasets, which demonstrates the effectiveness of our methods.

\section{Acknowledgements}
This paper was supported by National Key R$ \& $D Program of China (2020YFC1523202).

\section{Declarations}
\subsection{Conflict of interest}
The authors declare that they have no conflict of interest.
\subsection{Availability of data}
All data are available upon request and clearance for dispersal.
\subsection{Code availability}
All code is available upon request and clearance for dispersal.

%

\appendix
\section{Comparison with pairwise task interpolation}
	Let $ G $ denote the random variable which takes a uniform distribution on the indices of the tasks. The covariance matrix induced by the individual task interpolation (MTI, MTST, FM) is $ Cov(\mathbb{E}[X_i^{new,l} |G,X_i^l]) $. Thus in our method, the regularization which depends on the covariance can be larger , and then the generalization error bound can be lower.
	Furthermore, we also compare the proposed MTI with MLTI (ICLR2020) , and difference of them exists in the generator $ G $.
	
	For MLTI, the generation process of new tasks is: 
	\begin{equation}
		\begin{aligned}
			G_{1}: \lambda X_{i}^{l}+(1-\lambda) X_{j}^{l}, j \in\{1, \ldots, n\},
		\end{aligned}
	\end{equation}
	where $ \lambda \sim \operatorname{Beta}(\alpha, \beta).  $
	
	For MTI, the generation process of new tasks is: 
	\begin{equation}
		\begin{aligned}
			G_{2}: \lambda_{1} X_{i}^{l}+\lambda_{2} X_{a}^{l}+\lambda_{3} X_{b}^{l}, a, b \in\{1, \ldots, n\},
		\end{aligned}
	\end{equation}
	where $ \lambda \sim \operatorname{Dirichlet}(\boldsymbol{\gamma}).$
	
	The covariance matrixes are:
	\begin{equation}
		\begin{aligned}
			\operatorname{Cov}&\left(E\left[X_{i}^{n e w, l} \mid G_{1}, X_{i}^{l}\right]\right)\\
			&=\operatorname{Cov}\left(\lambda X_{i}^{l}+(1-\lambda) \sum_{j=1}^{n} X_{j}^{l}\right),\\
			\operatorname{Cov}&\left(E\left[X_{i}^{n e w}, l \mid G_{2}, X_{i}^{l}\right]\right)\\
			&=\operatorname{Cov}\left(\lambda_{1} X_{i}^{l}+\lambda_{2} \sum_{a=1}^{n} X_{a}^{l}+\lambda_{3} \sum_{b=1}^{n} X_{b}^{l}\right)\\
			&=\operatorname{Cov}\left(\lambda_{1} X_{i}^{l}+\left(1-\lambda_{1}\right) \sum_{j=1}^{n} X_{j}^{l}\right).
		\end{aligned}
	\end{equation}
	Given that $ \left\{X_{j}^{l}\right\}_{j=1}^J $ are the same in the above covariance matrixes, comparing them is equal to compare $ \operatorname{Cov}(\lambda) $ and $ \operatorname{Cov} (\lambda_1)$.
	Based on their definition, it is easy to get that:
	\begin{equation}
		\begin{aligned}
			\operatorname{Cov}(\lambda)=\frac{\alpha \beta}{(\alpha+\beta)^{2}(\alpha+\beta+1)} ,
		\end{aligned}
	\end{equation}
	\begin{equation}
		\begin{aligned}
			\operatorname{Cov}\left(\lambda_{1}\right)=\frac{\gamma_{1}\left(\widehat{\gamma}-\gamma_{1}\right)}{\widehat{\gamma}^{2}(\widehat{\gamma}+1)}, where~\hat{\gamma}=\sum_{j=1}^{n} \gamma_{j}.
		\end{aligned}
	\end{equation}
	
	When $  \alpha,\beta, \gamma <1 $, there can exist $ \operatorname{Cov}\left(\lambda\right)>\operatorname{Cov}\left(\lambda_{1}\right) $. It is worth noting that when $ \alpha=\beta=0.2 $, the model can usually achieve the best performance, and in this case we have $ \operatorname{Cov}\left(\lambda\right) \leq \operatorname{Cov}\left(\lambda_{1}\right)  $, which means that our MTST can achieve better generalization than MLTI.
	
\section{Proof of the Theorem}	
	The detailed derivation of Eq. \ref{proof1} is as follows.
	\begin{small} 
		\begin{equation}
			\begin{aligned}
				&\hat{\mathcal{R}}_{n}\left(\mathcal{F}_{R}\right) =\mathbb{E}_{\xi} \sup _{f \in \mathcal{F}_{R}} \frac{1}{n} \sum_{i=1}^{n} \xi_{i} f\left(X_{i}\right) \\
				&=\mathbb{E}_{\xi} \sup _{\|\theta\|_{\Sigma}^{2} \leq R} \frac{1}{n} \sum_{i=1}^{n} \xi_{i} \theta^{\top} X_{i} \\
				&=\mathbb{E}_{\xi} \sup _{\|\theta\|_{\Sigma}^{2} \leq R} \mathbb{E}_{X^{new}} \frac{1}{n} \sum_{i=1}^{n} \xi_{i} \theta^{\top}\left(X_{i}-X^{new}\right)\\
				&\leq \frac{\sqrt{R}}{n} \mathbb{E}_{X^{new}} \mathbb{E}_{\xi} \sqrt{\sum_{i=1}^{n} \sum_{j=1}^{n} \xi_{i} \xi_{j}\left(\Sigma^{\frac{\dagger}{2} }\left(X_{i}-X^{new}\right)\right)^{\top}\left(\Sigma^{\frac{\dagger}{2}}\left(X_{j}-X^{new}\right)\right)}\\
				&\leq \frac{\sqrt{R}}{n} \sqrt{\mathbb{E}_{X^{new}} \mathbb{E}_{\xi} \sum_{i=1}^{n} \sum_{j=1}^{n} \xi_{i} \xi_{j}\left(\Sigma^{\frac{\dagger}{2}}\left(X_{i}-X^{new}\right)\right)^{\top}\left(\Sigma^{\frac{\dagger}{2}}\left(X_{j}-X^{new}\right)\right)}\\
				&=\frac{\sqrt{R}}{n} \sqrt{\mathbb{E}_{X^{new}} \sum_{i=1}^{n}\left(\Sigma^{\frac{\dagger}{2}}\left(X_{i}-X^{new}\right)\right)^{\top}\left(\Sigma^{\frac{\dagger}{2}}\left(X_{i}-X^{new}\right)\right)}\\
				&=\frac{\sqrt{R}}{n} \sqrt{\mathbb{E}_{X^{new}} \sum_{i=1}^{n}\left(X_{i}-X^{new}\right)^{\top} \Sigma^{\dagger}\left(X_{i}-X^{new}\right)}.
			\end{aligned}
		\end{equation}
	\end{small}
	Here $ \Sigma^{\dagger} $ denotes the Moore–Penrose inverse of $ \Sigma $. 
	
	The detailed derivation of Eq. \ref{proof2} is as follows.
	\begin{small} 
		\begin{equation}
			\begin{aligned}
				&\mathcal{R}_{n}\left(\mathcal{F}_{R}\right)=\mathbb{E}_{X_{1}, \ldots, X_{n}} \hat{\mathcal{R}}_{n}\left(\mathcal{F}_{R}\right) \\
				& \leq \mathbb{E}_{X_{1}, \ldots, X_{n}} \frac{\sqrt{R}}{n} \sqrt{\sum_{i=1}^{n} \mathbb{E}_{X^{new}}\left(X_{i}-X^{new}\right)^{\top} \Sigma^{\dagger}\left(X_{i}-X^{new}\right)} \\
				& \leq \frac{\sqrt{R}}{n} \sqrt{\sum_{i=1}^{n} \mathbb{E}_{X_{i}, X^{new}}\left(X_{i}-X^{new}\right)^{\top} \Sigma^{\dagger}\left(X_{i}-X^{new}\right)} \\
				&=\frac{\sqrt{R}}{n} \sqrt{\sum_{i=1}^{n} \sum_{k, l}\left(\Sigma^{\dagger}\right)_{k l} \mathbb{E}_{X_{i}, X^{new}}\left(X_{i}-X^{new}\right)_{k}\left(X_{i}-X^{new}\right)_{l}} \\
				&=\frac{\sqrt{R}}{n} \sqrt{\sum_{i=1}^{n} \sum_{k, l}\left(\Sigma^{\dagger}\right)_{k l}(\Sigma)_{k l}} \\
				&=\frac{\sqrt{R}}{n} \sqrt{\sum_{i=1}^{n} \operatorname{tr}\left(\Sigma \Sigma^{\dagger}\right)}=\frac{\sqrt{R}}{n} \sqrt{\sum_{i=1}^{n} \operatorname{rank}(\Sigma)}\\
				&=\frac{\sqrt{R} \sqrt{\operatorname{rank}(\Sigma)}}{\sqrt{n}}.
			\end{aligned}
		\end{equation}
	\end{small}

\end{document}